%% file: main.tex
\documentclass{article} 
\usepackage{iclr2027_conference,times}

\input{math_commands.tex}

\usepackage{array}
\usepackage{longtable}
\usepackage{booktabs}
\usepackage{hyperref}
\usepackage{url}
\usepackage{listings}
\usepackage{comment}
\usepackage{float}
\usepackage{amsmath,amssymb,booktabs,multirow,enumitem,graphicx}
\usepackage{tikz}
\usetikzlibrary{positioning,arrows.meta,calc,decorations.pathreplacing}
\newcommand{\bench}{\textsc{STEMBench}}

\title{Benchmarking LLM abilities on postdoc-level natural science problems}

\author{Alina Avanesyan$^{1}$, Alexey Boldyrev$^{1}$, Liqun Deng$^{3}$, Denis Derkach$^{1}$, \\
\textbf{Mary Godunova$^{1}$, Mikhail Hushchyn$^{1}$, Dmitry Ilvovsky$^{1}$, Irina Knyazeva$^{2}$,} \\
\textbf{Irina Piontkovskaya$^{4}$, Dmitry Polubotko$^{4}$, Fedor Ratnikov$^{1}$,} \\
\textbf{Ekaterina Trofimova$^{5,6,1}$, Zhenhe Zhang$^{4}$} \\
$^{1}$HSE University, Moscow, Russia \\
$^{2}$N.~P.~Bechtereva Institute of the Human Brain RAS, St.~Petersburg, Russia \\
$^{3}$Independent researcher \quad
$^{4}$AI Foundation and Algorithm Lab\\
$^{5}$Constructor Labs \quad
$^{6}$Constructor University, Bremen, Germany
}

\iclrfinalcopy
\begin{document}

\maketitle

\begin{abstract}
Scientific problems have no answer key, so evaluating an agent means deciding
what counts as a better solution before any agent is run. Existing benchmarks
either match a symbolic answer against a key or, where they require executable
work, score it by unit tests, by a rubric, or by position on a competition
leaderboard. None of these is the criterion by which a field decides that one
published method beats another. \bench{} adopts that criterion directly. Each
of its twelve research problems in high-energy physics, astrophysics and
neuroscience is scored by its own field's figure of merit, and each comes with two measured anchors: a submission that uses no domain knowledge, and an
expert reference reproducing a published method, both obtained by running code
shipped with the benchmark. The interval between them states how much of a
problem is crossed by scientific insight rather than by engineering. Across eight frontier models, task-level wins and domain-level leadership
dissociate: a model that wins five of the twelve tasks leads no domain mean. In the high-effort cross-model comparison, at least one model's mean score exceeds the expert anchor on seven of twelve tasks. The remaining five span uncertainty estimation, latent and causal inference, and cross-site clinical classification. Numerical scores and official-submission validity capture different aspects of performance and are reported separately. We release the tasks, graders, anchors and harness.
\end{abstract}

\section{Introduction}
\label{sec:intro}
 
Asking whether an agent solved a scientific problem
requires deciding, before the agent is run, what makes one solution
better than another. That decision is the hard part of building an evaluation
for scientific work, and it is the part that existing benchmarks answer by
substituting a criterion that is easier to automate: a reference answer, a
test suite, a rubric, or a position in a competition field
(Section~\ref{sec:related}). None of these is the criterion by which a field
decides that one published method beats another.
 
We present \bench{}, twelve research problems in high-energy physics (HEP),
astrophysics, and neuroscience that adopt that criterion directly. Each problem
is scored by the figure of merit its own community already uses to compare
published methods, from background rejection at fixed signal efficiency to
coverage-penalised interval width. Each comes with two anchors obtained by
running code shipped with the benchmark: a floor produced by a submission that
uses no domain knowledge, and a ceiling produced by reproducing a published
method. The interval between them states how much of the problem is crossed by
a scientific insight rather than by engineering, and it is against that
interval, not against a perfect score, that we read what an agent achieves.
Agents work offline in a sandbox over staged data of $0.1$ to $12$~GB and are
graded host-side on withheld splits.
 
Our contributions are:
 
\begin{itemize}[leftmargin=1.2em, itemsep=1pt, topsep=2pt]
\item \textbf{A scoring design anchored at both ends.} Every task ships a
baseline and an expert reference submission, both runnable end to end on one
workstation, so the interval a solution must cross is measured rather than
assumed, and neither anchor is a number quoted from a paper
(Section~\ref{sec:formal}).
\item \textbf{Twelve task contracts with community metrics.} One uniform
layout across three domains, with ground truth held host-side, retrieval
restricted to a curated per-task corpus, and metrics carrying coverage
penalties and geometric gates as explicit defences against exploitation
(Section~\ref{sec:bench}).
\item \textbf{An evaluation of eight frontier models.}
Task-level wins and domain-level leadership dissociate. At least one
model's mean score exceeds the expert anchor on seven of twelve tasks
in the high-effort cross-model comparison. The remaining five span
uncertainty estimation, latent and causal inference, and cross-site
clinical classification. Official-submission validity is reported
separately from the numerical score (Section~\ref{sec:results}).
\end{itemize}

 
\section{Related Work}
\label{sec:related}

\begin{table}[ht!]
\caption{Selected related benchmarks and their evaluation targets. Rows
summarise the main assessment criterion, not the only one. \bench{}
reports its baseline and expert-reference scores alongside the task score.}
\label{tab:related}
\centering
\scriptsize
\setlength{\tabcolsep}{3pt}
\begin{tabular}{@{}p{0.23\textwidth}p{0.32\textwidth}p{0.39\textwidth}@{}}
\toprule
\textbf{Benchmark} & \textbf{Evaluation target} & \textbf{Assessment or performance reference} \\
\midrule
ScienceAgentBench & Data-driven scientific programs & Program and execution evaluation \\
Terminal-Bench-Science & Scientific workflows & Task-specific terminal verification \\
ScienceBoard & Scientific software workflows & Task-completion checks \\
LabOSBench & Scientific instrument control & Subtask completion within a step budget \\
MLE-bench & ML competition submissions & Competition metrics and human leaderboard results \\
BioML-bench & Biomedical ML pipelines & Task performance and human baselines \\
PRBench & Physics-paper reproduction & Reproduced quantitative results and scoring rubrics \\
MLR-Bench & Open-ended ML research & LLM reviewers and research-quality rubrics \\
COMPOSITE-STEM & Expert STEM tasks & Exact-match and rubric-based jury grading \\
LifeSciBench & Life-science research responses & Expert-authored task-specific rubrics \\
InnovatorBench & Runnable LLM-research artifacts & Correctness, performance, and output quality \\
NatureBench & Scientific ML solutions & Source-paper state-of-the-art performance \\
\midrule
\bench{} (ours) & Computational solutions in HEP, astrophysics, and neuroscience & Scientific task metrics; executable baseline and expert anchors \\
\bottomrule
\end{tabular}
\end{table}

\paragraph{Scientific knowledge and reasoning.}
Physics and mathematics benchmarks score an answer against a reference
\citep{xu2025ugphysics,qiu2025phybench,bieganowski2025putnamlike}, and
multimodal STEM evaluations and domain question-answering collections
extend this setting to visual evidence and specialised knowledge
\citep{shen2024vlstem,zaki2023mascqa,rein2023gpqa}. Agentic systems now
reach elite-human level here: Physics Supernova attains 23.5/30 on the
IPhO~2025 theory problems, placing 14th of 406 contestants
\citep{qiu2025supernova}. The criterion, however, is still a marking scheme
for problems with known solutions. These evaluations measure important
prerequisites for scientific work, but what they score is the quality of an
answer, not of a method evaluated on withheld data under a community figure
of merit.

\paragraph{Benchmarks that score executable work.}
SciCode scores scientific code by unit tests \citep{tian2024scicode},
ScienceAgentBench scores self-contained programs for 102 tasks drawn from 44
papers \citep{chen2025scienceagentbench}, and Terminal-Bench-Science verifies
expert-curated research workflows in a terminal, with one oracle solution per
task demonstrating feasibility \citep{Terminal-Bench-Science_2026}. MLE-bench
scores 75 Kaggle competitions, and its multilingual and biomedical extensions
and MLR-Bench carry the setting to other languages, to biomedical pipelines and
to open-ended ML research
\citep{chan2025mle,trofimova2026ml2b,miller2025biomlbench,chen2025mlrbench}.
These are close to the classification and anomaly-detection tasks of \bench{},
but not to its inversion and latent-parameter tasks, where predictive accuracy
alone does not describe scientific validity. PaperBench, CORE-Bench and
PRBench, the last with 30 tasks across 11 physics subfields, score the
reproduction of published results
\citep{starace2025paperbench,siegel2024core,qiu2026prbench}. ScienceBoard and LabOSBench move agents into scientific software and
instrument simulators, 169 tasks across four domains and 96 subtasks on
eight instruments, and count a task as solved when the required
operations are completed \citep{sun2026scienceboard,zou2026labosbench}. Reproduction
is also how we build expert anchors, although the evaluated agents
need not recover the reference method. A test suite, a
terminal check or a completion check \emph{accepts or rejects} a solution, whereas a field
\emph{orders} competing methods on a continuous figure of merit. \bench{} adopts
the ordering and scores it on a withheld split.

\paragraph{Scoring relative to a reference.}
MLE-bench reads scores against a human leaderboard, InnovatorBench against a
reference solution on 20 LLM-research tasks, and NatureBench against the
reported state of the art of the 90 Nature-family papers its tasks come from
\citep{wu2026innovatorbench,wang2026naturebench}. \bench{} ships two
anchors per task, both measured by running code in the repository: a
no-domain-knowledge floor and a reproduced published method. It reports raw
scores beside them rather than a single normalised quantity
(Section~\ref{sec:formal}), and treats the expert anchor as an empirical
reference, not an upper bound.

\paragraph{Autonomous discovery agents.}
A parallel line of work builds agents that run the discovery loop end to
end; a recent survey organises the field into process-, autonomy- and
mechanism-oriented views \citep{wei2025agentic}. Kosmos iterates
literature search, data analysis and hypothesis generation for up to
twelve hours per run and is validated by experts auditing its final
reports, with 79.4\% of statements judged accurate
\citep{mitchener2025kosmos}. Such assessment is made after the run, by
readers, and cannot be re-executed. \bench{} fixes an executable
acceptance criterion \emph{before} any agent runs, bounded by two measured
anchors, which makes it a regression test for the analysis stage of the
discovery loop rather than a competitor to end-to-end evaluation.

\paragraph{Expert-rubric evaluations.}
COMPOSITE-STEM combines exact-match assessment with criterion-based
rubrics and LLM-jury grading on 70 expert-written tasks, and LifeSciBench
assesses 750 life-science workflow tasks against expert-authored rubrics
\citep{waters2026composite,liu2026lifescibench}. Rubrics capture aspects
of scientific judgment that resist reduction to a number; \bench{}
concentrates on computational problems where an explicit, executable
scoring rule can order the submitted results. Table~\ref{tab:related}
summarises the comparison.

\section{Problem Formulation}
\label{sec:problem}
\label{sec:unique}

We call a problem \emph{scientific}, as opposed to an \emph{exercise}, when it
has the seven properties of Table~\ref{tab:unique}. Each one breaks an
assumption that current STEM benchmarks rely on, and each is exercised by at
least one task in \bench{}.

\begin{table}[h]
\caption{Seven properties that separate a scientific problem from an exercise,
each with an example from \bench{}.}
\label{tab:unique}
\centering
\small
\setlength{\tabcolsep}{4pt}
\renewcommand{\arraystretch}{1.15}
\begin{tabular}{@{}>{\raggedright\arraybackslash}p{0.26\textwidth}
                >{\raggedright\arraybackslash}p{0.70\textwidth}@{}}
\toprule
\textbf{Property} & \textbf{What holds instead of an answer key} \\
\midrule
\textbf{P1} Underdetermined solution set
  & distinct methods are all admissible and are \emph{ordered} by a continuous
    figure of merit, not accepted or rejected \\
\textbf{P2} Model and code as deliverable
  & an executable artefact that generalises to a withheld split; on two tasks
    the grader itself trains the submitted model \\
\textbf{P3} Load-bearing domain priors
  & the gap to the expert is closed by a domain insight, e.g.\ a decorrelation
    condition for weak supervision or a canonical basis for Stokes profiles \\
\textbf{P4} Data beyond the context
  & $0.1$--$12$~GB of staged data, reached through loaders and subsampling;
    deciding what to look at is part of the task \\
\textbf{P5} Community figure of merit
  & gated composites and non-monotone rules: a sharper interval that
    under-covers loses to a wider one that covers \\
\textbf{P6} Silent failures
  & plausible errors, such as site leakage or a forward model that fits the
    spectrum but breaks the physics, which only withheld data exposes and an
    LLM judge reading the output cannot see \\
\textbf{P7} Empirical ceiling
  & the literature ceiling for cross-scanner ASD classification is about
    $0.75$ AUC, so progress is distance to the best published solution,
    not to $1.0$ \\
\bottomrule
\end{tabular}
\end{table}

\paragraph{Tasks and scores.}
\label{sec:formal}
A task instance is a tuple
$\tau = \bigl(\mathcal{P}, \mathcal{D}_{\mathrm{tr}}, \mathcal{D}_{\mathrm{te}},
\mathcal{C}, \mu, b, e\bigr)$, where $\mathcal{P}$ is the natural-language
specification (motivation, input/output contract, data description, metric
description), $\mathcal{D}_{\mathrm{tr}}$ and $\mathcal{D}_{\mathrm{te}}$ are
the visible and withheld splits, $\mathcal{C}$ is the submission contract
accepted by the scorer, $\mu : \mathcal{C} \to [0,1]$ is a deterministic
scoring function, and $b, e$ are two anchor scores defined below.

An agent is a policy $\pi$ that interacts with a sandboxed environment under a
budget $B$ (wall-clock, tokens, messages) and returns a submission
$\sigma_\pi \in \mathcal{C}$. Two anchors are shipped with every task:
a \emph{baseline} $b = \mu(\sigma_{\mathrm{base}})$, obtained by a submission
that uses no domain knowledge, and an \emph{expert reference}
$e = \mu(\sigma_{\mathrm{ref}})$, obtained by reproducing a published method.
Both are organiser-side artefacts. They are never staged into the agent
sandbox and play no role in an episode beyond fixing the scale.

The harness reports three quantities per episode: the raw task score
$\mu(\sigma_\pi)$ in the task's native definition, the gap above the baseline
\begin{equation}
\label{eq:delta}
\Delta_{\mathrm{base}}(\pi, \tau) \;=\; 100 \cdot \bigl(\mu(\sigma_\pi) - b\bigr)
\quad \text{[percentage points]},
\end{equation}
and \texttt{official\_submission\_valid}, a reproducibility gate that is
independent of the numeric score and requires a syntactically valid entry
point, a README with a runnable command, and every declared reproducibility
file. We report raw scores with both anchors alongside, because the quantity
the benchmark is designed around is the headroom
\begin{equation}
\label{eq:headroom}
h(\tau) \;=\; e - b ,
\end{equation}
the part of the task that is crossed by a scientific insight rather than by
engineering. A normalised score
$\hat{\mu} = \mathrm{clip}_{[0,1]}\bigl((\mu(\sigma_\pi) - b)/h(\tau)\bigr)$
can be derived post hoc from the reported triple, but we do not use it as the
headline number: it discards information whenever $h(\tau)$ is narrow, and on
this task set the narrow-headroom cases are exactly the scientifically hardest
ones.

\section{The \bench{} Benchmark}
\label{sec:bench}

\definecolor{taskC}{HTML}{0072B2}   
\definecolor{agentC}{HTML}{E69F00}  
\definecolor{gradeC}{HTML}{009E73}  
\definecolor{hepC}{HTML}{56B4E9}    
\definecolor{astC}{HTML}{CC79A7}    
\definecolor{neuC}{HTML}{D55E00}    

\begin{figure}[h]
\centering
\resizebox{\textwidth}{!}{%
\begin{tikzpicture}[
    font=\small,
    box/.style  ={draw, rounded corners=2pt, align=left, inner sep=4pt, line width=0.7pt},
    pkg/.style  ={box, draw=taskC,  fill=taskC!8},
    ag/.style   ={box, draw=agentC, fill=agentC!10},
    gr/.style   ={box, draw=gradeC, fill=gradeC!9},
    dom/.style  ={draw, rounded corners=2pt, align=center, inner sep=2pt,
                  minimum width=16mm, minimum height=5.5mm, line width=0.6pt},
    arr/.style  ={-{Latex[length=2mm]}, semithick},
    lab/.style  ={font=\scriptsize, align=center}
]

\node[pkg] (task) at (0,0) {%
  \textbf{\color{taskC!80!black}Task contract $\tau$}\\[1pt]
  \texttt{config.yaml}\\
  \texttt{description.md}\\
  \texttt{metric\_description.md}\\
  \texttt{dataset/public/},\ \texttt{private/}\\
  \texttt{references/} (8 of 12)\\
  \texttt{baseline\_solution/}\,$\to b$\\
  \texttt{expert\_solution/}\,$\to e$
};

\node[dom, draw=hepC!85!black, fill=hepC!20] (hep) at ([xshift=-19mm, yshift=10mm]task.north) {HEP $\times 4$};
\node[dom, draw=astC!85!black, fill=astC!20] (ast) at ([xshift=  0mm, yshift=10mm]task.north) {Astro $\times 3$};
\node[dom, draw=neuC!85!black, fill=neuC!18] (neu) at ([xshift= 19mm, yshift=10mm]task.north) {Neuro $\times 5$};
\draw[arr, hepC!85!black] (hep.south) -- ([xshift=-11mm]task.north);
\draw[arr, astC!85!black] (ast.south) -- (task.north);
\draw[arr, neuC!85!black] (neu.south) -- ([xshift= 11mm]task.north);

\node[ag, right=15mm of task] (agent) {%
  \textbf{\color{agentC!55!black}Agent $\pi$}, budget $B$\\[1pt]
  validation-gated ReAct loop\\
  \texttt{bash}, \texttt{python}, \texttt{data\_explorer}\\
  \texttt{local\_library\_search}, \texttt{document\_reader}\\
  \texttt{validation\_feedback} (no score)\\
  offline sandbox, no network
};

\node[gr, right=15mm of agent] (eval) {%
  \textbf{\color{gradeC!75!black}\texttt{grade.py}} (host-side)\\[1pt]
  withheld $\mathcal{D}_{\mathrm{te}}$\\
  $\mu(\cdot)\in[0,1]$\\
  \texttt{official\_submission\_valid}\\
  authored independently\\
  of \texttt{expert\_solution/}
};

\draw[arr, taskC!80!black] (task) -- node[lab, above]{$\mathcal{P},\ \mathcal{D}_{\mathrm{tr}}$}
                    node[lab, below]{read-only} (agent);
\draw[arr, agentC!60!black] (agent) -- node[lab, above]{$\sigma_\pi \in \mathcal{C}$}
                     node[lab, below]{submission} (eval);
\draw[arr, dashed, agentC!60!black] (agent.south) to[out=-90, in=-90, looseness=0.7]
      node[lab, below, pos=0.5, yshift=-0.5mm]{local runs, self-validation}
      (task.south east);

\coordinate (a0) at ($(task.south west)+(0,-21mm)$);
\coordinate (a1) at ($(eval.south east |- task.south)+(0,-21mm)$);
\fill[gradeC!15] ($(a0)!0.20!(a1)+(0,-0.9mm)$) rectangle ($(a0)!0.82!(a1)+(0,0.9mm)$);
\draw[semithick, black!70] (a0) -- (a1);
\foreach \p/\l/\c in {0/{$0$}/black!70, 0.20/{$b$}/black!55, 0.55/{$\mu(\sigma_\pi)$}/agentC!60!black,
                      0.82/{$e$}/gradeC!75!black, 1/{$1$}/black!70}{%
  \draw[semithick, \c] ($(a0)!\p!(a1)$) -- ++(0,-1.4mm)
        node[below, font=\scriptsize, text=\c]{\l};
}
\fill[agentC!65!black] ($(a0)!0.55!(a1)$) circle (1.4pt);
\node[lab, above right] at ([yshift=2.2mm]a0) {raw score $\mu$ and reported gap $\Delta_{\mathrm{base}}$};
\draw[decorate, decoration={brace, amplitude=3pt}, gradeC!75!black, semithick]
      ($(a0)!0.20!(a1)+(0,1.5mm)$) -- ($(a0)!0.82!(a1)+(0,1.5mm)$)
      node[midway, above, font=\scriptsize, yshift=2pt, text=gradeC!65!black]
      {scientific headroom $h = e-b$};

\end{tikzpicture}}
\caption{The \bench{} evaluation loop. The agent reads the task specification
and the public split, works in an offline sandbox through six tools, and returns
a submission in the task's contract. Ground truth never enters the sandbox: the
grader runs host-side on the withheld split and returns $\mu\in[0,1]$ with a
reproducibility verdict, read against the anchors $b$ and $e$.}
\label{fig:overview}
\end{figure}
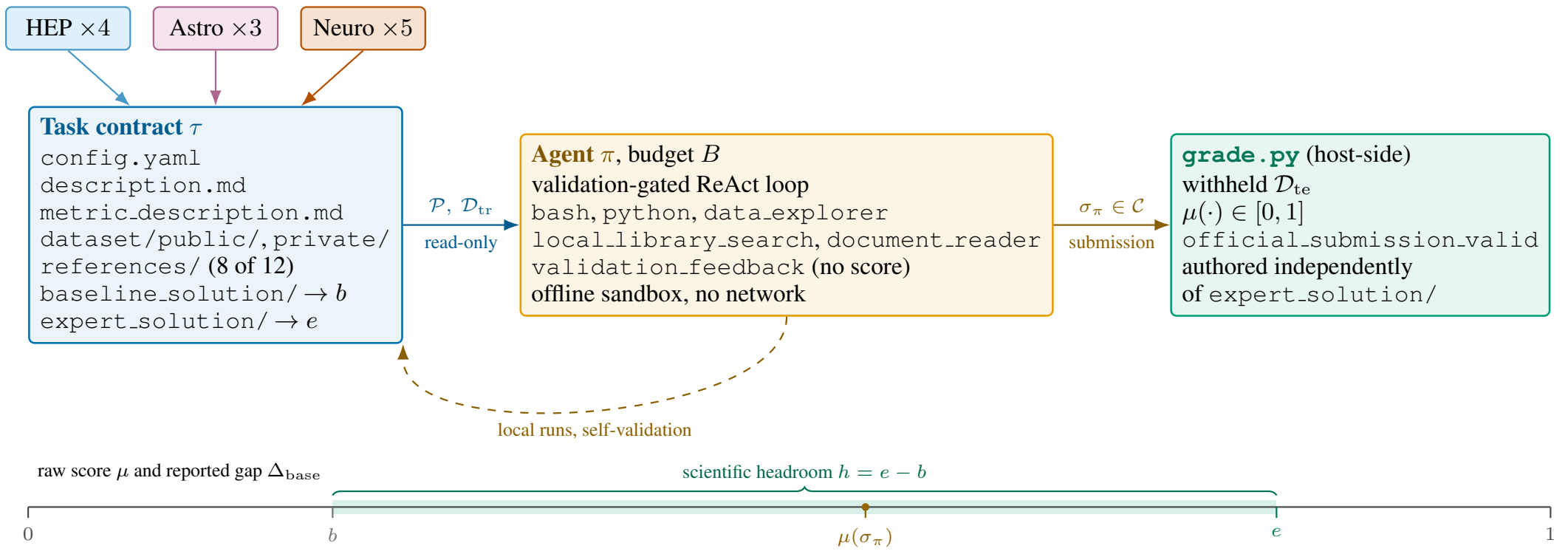

\bench{} consists of twelve research problems in three domains: four in
high-energy physics, three in astrophysics and five in neuroscience. Every
problem ships as a self-contained contract folder with a task description, a
metric specification, a role-tagged data manifest, a declared submission
contract, a host-side grader, and the two anchor submissions.
Figure~\ref{fig:overview} shows the resulting evaluation loop and
Table~\ref{tab:tasks} lists the tasks with their anchors.
 
The three domains were selected because they stress different parts of the
capability: high-energy physics contributes large-scale data and
hypothesis-free search, astrophysics contributes ill-posed inversion from
remote observation with uncertainty quantification, and neuroscience
contributes multi-scale forward models and cross-site generalisation on real
clinical data.
 
\subsection{Task Selection}
\label{sec:tasks}
 
Each of the twelve problems satisfies five criteria, applied before any solution
was written. The problem falls within the scientific scope of the domain; a
solution by classical or machine-learning methods is known to exist, so the
task is solvable rather than open; a typical solution requires several
non-trivial reasoning steps; a clear mathematical metric is definable on the
submission; and the problem exercises a metric family that the others do not.
The last criterion is what prevents a single agent from topping the aggregate
by specialising in one scoring rule.
 
The neuroscience block is the one place where we revised the selection after
the first round of runs. A single composite neural-mass-model task bundled
three distinct inferential operations, model-family identification from
generated dynamics, causal digital-twin response prediction, and latent
parameter inference, into one weighted score, so a high score on one operation
could mask a failure on another. The three now stand as independent contracts
with their own hidden-target boundaries, submission schemas and anchors, and
the legacy composite is retired.


Table~\ref{tab:tasks} lists the twelve problems with the unit each metric is
computed over, the family its scoring rule belongs to, and the two anchors.
Appendix~\ref{app:tasks} states every task in the terms the agent receives it,
and Appendix~\ref{app:prompts} reproduces one complete agent-facing prompt.
 
\begin{table}[t]
\caption{The twelve \bench{} tasks. ``Eval.\ units'' is what the metric is
computed over once. Anchors $b \to e$ are the \texttt{reference\_scores}
declared in each task's \texttt{config.yaml}, obtained by running the shipped
baseline and expert submissions end to end on a single workstation.}
\label{tab:tasks}
\centering
\small
\setlength{\tabcolsep}{3pt}
\resizebox{\textwidth}{!}{%
\begin{tabular}{@{}l l r l c@{}}
\toprule
\textbf{Domain} & \textbf{Task} & \textbf{Eval.\ units} &
\textbf{Metric family} & \textbf{$b \to e$} \\
\midrule
\multirow{4}{*}{HEP}
 & New-physics search (LHCO)      & 135{,}261 events   & ROC-AUC, SIC-AUC, rejection        & $0.327 \to 0.912$ \\
 & Higgs uncertainty (FAIR Univ.) & 100 pseudo-exps.   & coverage-penalised interval        & $0.400 \to 0.687$ \\
 & Top tagging (TopLandscape)     & 403{,}999 jets     & ROC-AUC, acc., rejection at $\epsilon_S$ & $0.063 \to 0.884$ \\
 & Dark Machines                  & 34 signal cells    & $\epsilon_S$ at fixed $\epsilon_B$, AUC & $0.003 \to 0.221$ \\
\midrule
\multirow{3}{*}{Astro}
 & Stokes inversion, simulated     & 10{,}000 spectra  & pinball loss $+$ forward gate      & $0.009 \to 0.624$ \\
 & Stokes inversion, real (Hinode) & 2 rasters         & $+$ spatial-structure term         & $0.000 \to 0.598$ \\
 & SN~Ia classification (PLAsTiCC) & 317 objects       & ROC-AUC, PR-AUC, log-loss          & $0.421 \to 0.849$ \\
\midrule
\multirow{5}{*}{Neuro}
 & ASD classification (ABIDE)      & 332 subjects      & held-out-site AUC, Brier, age      & $0.000 \to 0.312$ \\
 & NMM family id.                  & 45 series         & chance-corrected accuracy          & $0.000 \to 0.900$ \\
 & NMM digital twins               & 12 interventions  & signed causal-response composite   & $0.100 \to 0.715$ \\
 & NMM parameter recovery          & 20 subjects       & median relative parameter error    & $0.125 \to 0.515$ \\
 & BOLD deconvolution              & 21 cases          & neural drive $+$ two kernel banks  & $0.012 \to 0.353$ \\
\bottomrule
\end{tabular}}
\end{table}

\subsection{The Uniform Task Contract}
 
Heterogeneity across scientific domains is real, but it should not be part of
what a benchmark measures. Every task is therefore organised in one
layout. A single \texttt{config.yaml} is the authoritative contract: it declares
the data manifest with a role per file (\texttt{training}, \texttt{test\_input},
\texttt{auxiliary}, \texttt{ground\_truth}), the required submission files with
the format checks applied to them, the host-side scorer entry point, the two
anchors, the package requirements and the compute requirements. Alongside it,
\texttt{description.md} carries the scientific specification handed to the
agent, \texttt{metric\_description.md} pins down the scoring rule,
\texttt{grade.py} implements it, \texttt{checksums.yaml} fixes the data, and
\texttt{baseline\_solution/} and \texttt{expert\_solution/} hold the two anchor
submissions with their own instructions for reproducibility. Eight tasks, all of
astrophysics and neuroscience, additionally ship a curated \texttt{references/}
corpus of papers and domain notes.
 
Only files tagged with a public role are staged into the sandbox. Ground-truth
files never leave the host and are read only by the grader after the episode
ends. Every active task exposes the same grading signature,
\texttt{grade(submission\_dir, answers\_path)}, which is what makes twelve
heterogeneous scientific workflows scoreable through one harness.
 
For the high-energy-physics family the upstream datasets are hosted externally,
so the repository bundles loaders, manifests and checksums while the bulk data
is fetched from the cited sources and preprocessed into the staged splits, which
range from $0.1$ to $12$~GB per task. All datasets used are openly accessible
with redistribution rights compatible with a public benchmark: Zenodo releases
for the LHC Olympics R\&D dataset and PLAsTiCC, public challenge releases for
TopLandscape and the FAIR Universe Higgs data, a public data-use agreement for
ABIDE, open solar data for the Hinode/SOT-SP rasters, and in-project generation
for the synthetic neuroscience tasks.
 
\subsection{Metrics and Defences Against Gaming}
 
Each of the twelve metrics instantiates the figure of merit by which the
corresponding community already compares published methods: SIC-AUC as
used to rank the LHC Olympics teams, the official coverage-and-width scoring
rule of the NeurIPS~2024 FAIR Universe competition, background rejection at
fixed signal efficiency as used in every modern jet-tagging paper, signal
efficiency at fixed background efficiency per (channel, signal) cell for
Dark Machines, the PLAsTiCC ranking metric, the axes along which solar
inversion codes are compared, the validation metrics of the Virtual Brain and
Balloon model papers, the rsHRF benchmark metrics, and macro AUC over held-out
acquisition sites for ABIDE. Each metric therefore inherits a literature floor
and ceiling, which is what makes the anchors in Table~\ref{tab:tasks}
interpretable.
 
Three failure modes are addressed by construction. Against \emph{metric
exploitation}, composite scores bound every component to $[0,1]$ before convex
combination, so no submission can inflate the headline by driving a single
component to an extreme, and the Higgs task applies an asymmetric coverage
penalty so that an under-covering interval cannot win on width alone. Against \emph{component substitution}, the two inversion tasks add a gate
of our own on top of the community metric: the forward-consistency and
spatial-structure terms enter as $\sqrt{R\,F}$ and $\sqrt{R\,S}$, so that a
submission cannot win on spectral reconstruction while failing to recover
parameters, and a declared forward model that is inconsistent with the
physical law it claims to implement is caught by a dedicated validity
battery and zeroes the corresponding term. Against
\emph{reward hacking on the reference}, every expert submission was produced by
reproducing a published method without consulting the scorer implementation, so
the metric and the reference were authored independently; a leakage attack would
surface as a discrepancy between the reference score and the published score of
the same method on the same task.
 
\subsection{Anchors}
 
The baseline anchor is deliberately weak but not degenerate: it is what a
capable engineer produces without domain knowledge, for example a constant
train-marginal quantile predictor on the inversion tasks, a range-midpoint
parameter estimate on neural-mass parameter recovery, a constant-$\mu$ interval
on the Higgs task, or a literal zero-response prediction on the digital-twin
task. Several metrics normalise against exactly that predictor, which is why
six baselines sit at or near $0.000$ by construction rather than by accident.
 
The expert anchor reproduces a published method end to end. The resulting gaps
are wide and scientifically meaningful, from $0.000 \to 0.900$ on neural-mass
family identification to $0.009 \to 0.624$ on simulated Stokes inversion, with
the deliberately narrow $0.000 \to 0.312$ on ABIDE reflecting the hard ceiling
that real cross-scanner clinical data imposes (P7). In every case the gap
was crossed by a domain-specific insight rather than by scaling compute or data,
which is the property that makes the headroom a measurement of scientific
reasoning rather than of engineering throughput.
 
\subsection{Environment and Toolchain}

The harness is built on Inspect-AI~\citep{UK_AI_Security_Institute_Inspect_AI_Framework_2024}. Publication runs use per-domain container images that mount only the public files declared in the manifest, read-only, with networking disabled and every
package pre-installed; Appendix~\ref{app:runconfig} lists the images, the hardware and the per-episode limits.

The agent-facing surface is six tools, fixed across tasks. \texttt{bash} and
\texttt{python} execute code in the sandbox. \texttt{data\_explorer} returns a
structural summary of a public data file (shape, dtype, statistics) without
loading it into the model context, over eight formats, and refuses
pickle-like extensions rather than implicitly unpickling them.
\texttt{local\_library\_search} and \texttt{document\_reader} expose the
task-local \texttt{references/} corpus, including PDF extraction, on the eight
tasks that ship one. \texttt{validation\_feedback} is a pre-submission checker
that reports missing files and schema violations and never reveals the numeric
score. There is no internet access, which would turn problems with published solutions into
retrieval exercises.
 
All results reported here use the \textbf{basic} agent; the \textbf{basic+references} variant is released with the harness and its evaluation is left to future work. The \textbf{basic} agent runs a
validation-gated ReAct loop in which a submission is accepted only after the
structural validator passes, with rejections returned as JSON feedback that the model can act on. The \textbf{basic+references} agent is identical except that the two retrieval tools are enabled where a corpus exists. Both are benchmark-pure: no hidden score, anchor or ground truth is revealed at any point. Each episode runs under one hour of wall-clock time, $10^6$ model tokens and
$200$ messages.

The system prompt is identical across tasks and models and is reproduced in
Appendix~\ref{app:prompts}. Two of its instructions affect how a score should be
read. The agent is told to produce a valid submission early and keep it valid,
so that a budget limit truncates improvement rather than erasing all work; an
episode that still ends with no valid submission is therefore a substantive
failure and not a formatting accident. The agent is also warned away from
constant, group-midpoint and globally broadcast predictions. Several baseline
anchors are exactly such predictors, so the scaffold tells the agent that the
trivial solution is known to be trivial, without disclosing its score.
 
\subsection{Evaluation Protocol}

Holding the harness fixed and varying the model isolates the core LLM; holding
the model fixed and varying the harness isolates the tooling. Both axes are
needed, since a gain reported on one is routinely produced by the other. The
harness passes a functional test suite and every contract passes checksum and
anchor-grading smoke tests (Appendix~\ref{app:runconfig}). A blinded
expert-agreement study, comparing the automatic ranking of intermediate
submissions between $b$ and $e$ with rankings by domain reviewers against
targets of $\tau \geq 0.85$ and $\geq 90\%$ pairwise agreement, is the
remaining validation step.

\section{Results}
\label{sec:results}

\begin{figure}[h!]
\centering
\includegraphics[width=\textwidth]{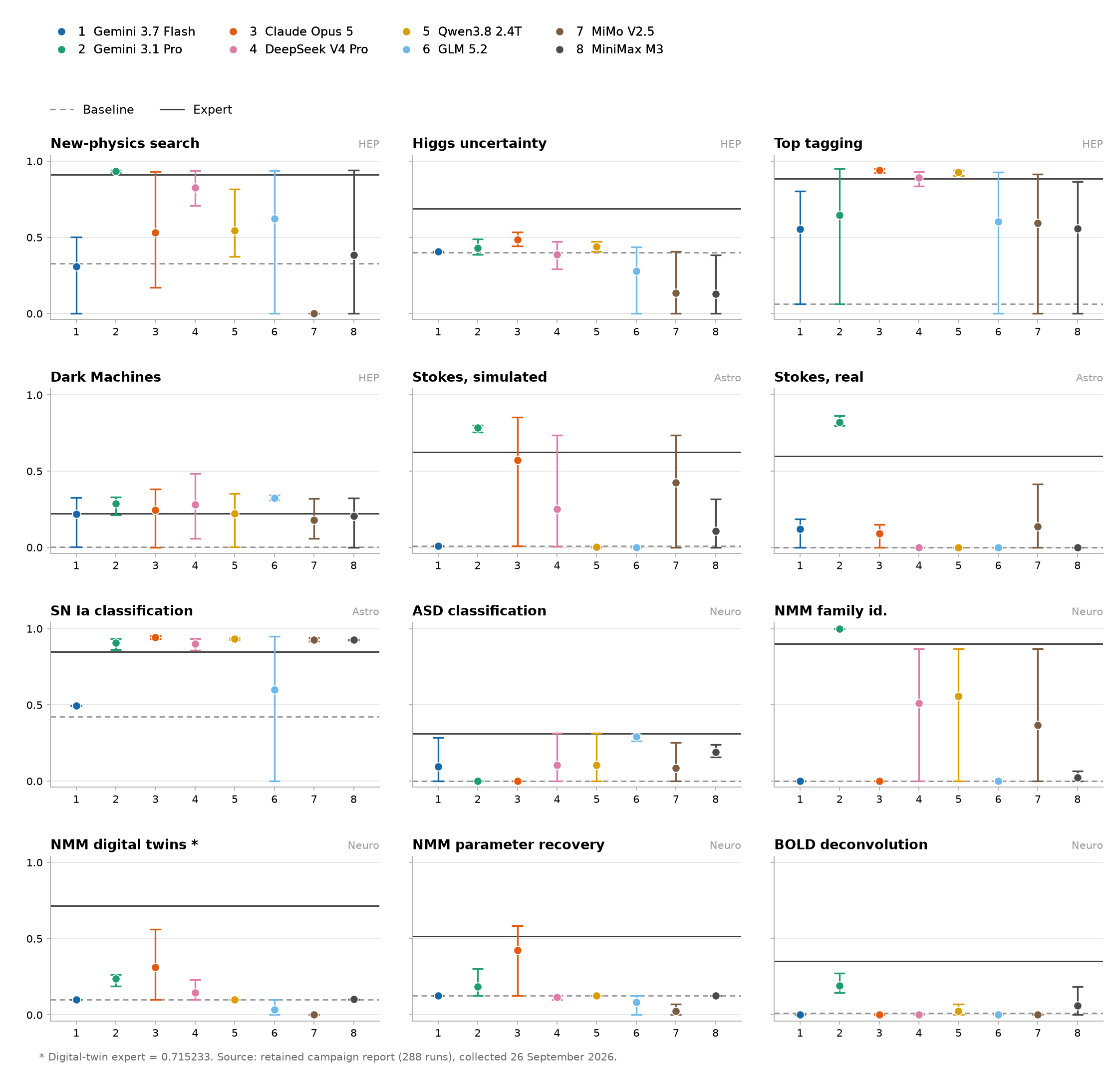}
\caption{Raw task scores of eight frontier models using the
\textbf{basic} agent at high reasoning effort. Each marker is the mean
of the three retained scored runs,
including zeros. Vertical bars show the minimum and maximum
of these three scores, not standard deviations or confidence
intervals. Horizontal lines mark the baseline $b$ (dashed) and expert
reference $e$ (solid). A numerical zero does not by itself identify
an invalid submission. The digital-twin expert anchor is $e=0.715233$.}
\label{fig:results}
\end{figure}

Figure~\ref{fig:results} reports eight frontier models under the
\textbf{basic} scaffold at high reasoning effort (exact values and domain means are in Appendix~\ref{app:results}). A \emph{retained} run is one whose model dialogue completed without infrastructure interruption. All retained runs enter every mean, including runs scoring zero.

Three observations follow. First, task-level wins and domain-level
leadership differ. DeepSeek~V4~Pro has the highest HEP mean ($0.60$),
while Gemini~3.1~Pro leads astrophysics ($0.84$) and neuroscience
($0.32$). Opus~5 and Gemini~3.1~Pro each lead five of the twelve
tasks; GLM~5.2 leads the remaining two. Opus nevertheless leads no
domain mean. In HEP, its wins on Higgs uncertainty and top tagging
coexist with a lower new-physics-search score, yielding a domain mean
of $0.55$, below DeepSeek's $0.60$. In neuroscience, its wins on
digital twins and parameter recovery coexist with means displayed
as $0.00$ on the other three tasks.

Gemini's domain means range from $0.32$ to $0.84$, a spread of $0.52$.
The cross-model spreads of domain means are $0.37$ in HEP, $0.64$
in astrophysics, and $0.26$ in neuroscience, using the displayed
values. Thus, this within-model spread exceeds the cross-model
spread in HEP and neuroscience, but not in astrophysics. A descriptive two-way sum-of-squares decomposition of the 96
unrounded task--model means assigns $60.5\%$ of the total variation
to task main effects, $11.5\%$ to model main effects, and the
remaining $28.0\%$ to departures from additivity. This residual
includes both task - model interaction and sampling variability
in the three-run cell means.

Second, at least one model's mean score exceeds the expert anchor
on seven of the twelve tasks: new-physics search, top tagging,
Dark Machines, both Stokes inversions, SN~Ia classification, and
neural-mass family identification. All model means remain below
the expert anchor on the other five: Higgs uncertainty (best mean
$0.48$, $e=0.687$), ASD classification on ABIDE ($0.29$, $e=0.312$),
neural-mass digital twins ($0.31$, $e=0.715233$), neural-mass parameter
recovery ($0.42$, $e=0.515$), and BOLD deconvolution ($0.19$, $e=0.353$).
On digital twins, the best mean remains closer to the baseline
($b=0.100$) than to the expert anchor. Aggregate scores
alone do not establish a single cause for these gaps. Counting individual runs rather than means, nine tasks exceed the expert anchor: ABIDE reaches $0.314$ and parameter recovery $0.583$.

Third, numerical scores and official-submission validity are distinct
outcomes. Of the 288 retained runs, 168 have an affirmative
official-submission flag, 64 a negative flag, and 56 an unavailable
flag. Among the 88 runs scoring exactly zero, these counts are 15,
17, and 56, respectively. Zero scores therefore cannot be interpreted
automatically as failures to produce an official submission.

Four model means on real Stokes are exactly zero, yet all twelve underlying
flags are unavailable; two ABIDE means round to $0.00$ but are positive,
with negative flags for Opus and affirmative ones for Gemini~3.1~Pro. Exposure to the wall-clock cap varies across models in the retained
campaign. It is recorded for 14 of 36 MiMo~V2.5 runs ($38.9\%$),
15 of 36 DeepSeek~V4~Pro runs ($41.7\%$), and 2 of 36
Gemini~3.7~Flash runs ($5.6\%$). These frequencies describe observed
stopping conditions; they do not quantify the score improvement
that a larger budget would produce. Because the cap is wall-clock, it also absorbs provider latency, so these runs measure capability under a fixed time budget rather than per-token capability.

\paragraph{Pilot effort sweep.} Before the campaign, a pilot sweep ran
Gemini~3.1~Pro at \texttt{low}, \texttt{medium} and \texttt{high}
reasoning effort on all twelve tasks (Appendix~\ref{sec:effort}). It used
an earlier task revision and predates a fix to GPU access in the
containers, so its absolute scores are not comparable with
Figure~\ref{fig:results}. Qualitatively, most of the gain came between
\texttt{low} and \texttt{medium}, several tasks were bimodal, with runs
landing either near the best score or at zero, and the effect of effort
was not monotone on every task. These observations motivated fixing the
campaign at \texttt{high} effort and repeating every cell three times.

\section{Conclusion}
\bench{} scores scientific problem solving using community metrics
and two measured anchors: a baseline and a reproduced expert reference.

In the high-effort comparison of eight frontier models, at least one
model's mean score exceeds the expert anchor on seven of twelve tasks,
while on others, such as digital-twin response prediction and BOLD
deconvolution, the best model remains far below it. The five remaining
gaps span uncertainty estimation, latent and causal inference, and
cross-site clinical generalisation. Task wins and domain leadership
differ, and numerical scores must be distinguished from
official-submission validity. This range of difficulty lets \bench{}
differentiate models by domain and task type rather than by a single
aggregate score.

These conclusions have limits.
Stable model rankings will require more runs per cell and confidence
intervals;  a pilot effort sweep (Appendix~\ref{sec:effort}) covers one model
on an earlier task revision; and agreement between the automatic metrics and blinded
expert rankings remains to be measured. 

We release the tasks, graders,
anchors and harness
\citep{anonymous_2026_22959834,anonymous_2026_22960135}.

\subsubsection*{Reproducibility Statement}
 
Every claim in this paper is tied to an artefact in the benchmark repository.
The twelve task contracts, their graders, both anchor submissions per task and
the harness are released; Section~\ref{sec:bench} describes the contract layout
and Appendix~\ref{app:prompts} reproduces the agent-facing prompt surface
verbatim. Anchors quoted in Table~\ref{tab:tasks} are the
\texttt{reference\_scores} in each task's \texttt{config.yaml} and are
reproduced by running the shipped baseline and expert submissions. Appendix~\ref{app:results} reports aggregate per-model scores for the
cross-model comparison. Appendix~\ref{sec:effort} reports a pilot effort and budget sweep on an earlier task revision. Data provenance and licences are given in
Section~\ref{sec:bench}, and per-domain container images pin the software
environment. The task
contracts, graders, both anchor submissions and staged data are
archived at \citet{anonymous_2026_22960135}, and the evaluation harness at
\citet{anonymous_2026_22959834}.
 
\subsubsection*{Ethics Statement}
 
The ABIDE task uses de-identified human neuroimaging data obtained under the
public ABIDE data-use agreement; no new human-subject data was collected and no
participant is identifiable from the staged files. The task predicts a clinical
label from resting-state fMRI, and we state explicitly in the task description
and in Appendix~\ref{app:tasks} that cross-scanner transfer of such a marker is a
known hard limit: the benchmark measures methodological competence on a public
research dataset and the scores reported here carry no clinical validity and
must not be read as evidence that any method is fit for diagnostic use. The
neural-mass digital-twin task is a controlled synthetic benchmark and is
likewise not a patient-specific model. All other datasets are simulated or
openly released physical-science data.
 
\subsubsection*{AI Use Disclosure}
 
We used generative AI tools in the preparation of this paper. Specifically: to
draft Sections~\ref{sec:problem} and~\ref{sec:bench} from internal project
reports authored by us; to develop the conceptual framing of those sections,
including the taxonomy of properties P1--P7 and the
formalisation of a task instance with its two anchors; to search for and verify
related benchmarks; to aggregate per-run scores into
Tables~\ref{tab:agg} and~\ref{fig:results} and to draft the accompanying
interpretation; and to edit prose for readability throughout. We did not use generative AI tools to formulate or prove mathematical claims, to
generate any benchmark data, or to write any task, grader, baseline or expert
solution in the benchmark itself. The twelve problems, their metrics, the
reference implementations and all experimental runs are the authors' own work.
All AI-assisted output was reviewed by the authors. The cross-model task and domain means in Table~\ref{tab:results}
and Figure~\ref{fig:results} were recomputed from the 288 retained
per-run scores in the campaign report. We take
responsibility for the final content of this work, including text, claims and
artifacts produced with the aid of generative AI. The language models evaluated in Section~\ref{sec:results} and Appendix~\ref{sec:effort}
are the object of study, not tools used to produce this paper.

\bibliography{iclr2025_conference}
\bibliographystyle{iclr2027_conference}
 
\appendix

\section{Task Descriptions}
\label{app:tasks}

This appendix states each of the twelve problems as the agent receives it:
what is visible in the sandbox, what the submission contract requires, and the
property that keeps a competent generic pipeline well below the expert anchor.
Evaluation units, metric families and the two anchors are given once, in
Table~\ref{tab:tasks}, and are not repeated here. Identifiers are the directory
names in the release.

\subsection{High-energy physics}

\paragraph{New-physics search (\texttt{lhco\_anomaly}).}
Half a million unlabelled dijet events, seven observables per jet; submit one
anomaly score per test event. No labels exist at any point, so the discriminant
must come from the unlabelled mixture itself at $S/B \sim 10^{-2}$. The
resonance masses are disclosed, as in the public LHC Olympics release, so the
difficulty is statistical rather than one of guessing the signal model \cite{kasieczka2021lhc}.

\paragraph{Higgs uncertainty (\texttt{higgsml\_uncertainty}).}
Ten million weighted simulated events; submit a point estimate of the signal
strength and a central interval per pseudo-experiment. Each pseudo-experiment
is drawn at its own hidden values of six nuisance parameters, so the test
distribution is shifted. Only coverage and width are scored: a sharp interval
that under-covers loses to a wider one that covers \cite{bhimji2026fair}.

\paragraph{Top tagging (\texttt{top\_quark\_tagging}).}
Jets as unordered variable-length sets of constituent four-momenta plus global
kinematics; submit a top-jet probability. Two of the four score components are
background rejections on a logarithmic scale, so a percent-level gain in
discrimination moves the headline far more than the same gain in accuracy.
Rewards permutation-invariant pooling, penalises flattening \cite{qu2022particle}.

\paragraph{Dark Machines (\texttt{darkmachines\_anomaly}).}
Four channels of background-only events, up to $7.7$ million in a ragged text
format. The submission is a model class with \texttt{fit} and \texttt{predict}
rather than a prediction file: the grader instantiates a fresh model per
channel, trains it on that channel's background and applies it to hidden files
holding every BSM signal. No signal is visible to the agent and the metric
cannot be computed locally \cite{aarrestad2021dark}.

\subsection{Astrophysics}

\paragraph{Stokes inversion, simulated (\texttt{stokes\_inversion}).}
Simulated Stokes spectra around one Fe~I line; submit three quantile arrays
over eleven Milne--Eddington parameters \emph{and} your own Unno--Rachkovsky
forward model \citep{deltoroiniesta2016inversion, mistryukova2023stokes}.
Azimuth is identifiable only modulo $180^\circ$, exact singular points make
some parameters unobservable, and filling factor trades against field strength.
The grader resynthesises the spectrum from the submitted parameters in an
isolated container.

\paragraph{Stokes inversion, real (\texttt{hinode\_inversion}).}
Real Hinode/SOT-SP rasters at a two-line pair; the same submission plus
complete parameter maps
\citep{deltoroiniesta2016inversion, mistryukova2023stokes}. The reference is a
Merlin inversion that is itself noisy, real instrument effects are present, and
the source-function convention needs a gain the forward model must apply. The
spatial term is scored on the parameter maps rather than on synthesised
spectra, so smoothing the output cannot buy it.

\paragraph{SN~Ia classification (\texttt{lightcurve\_classification}).}
Irregularly sampled, heteroscedastic six-band light curves with fewer than five
hundred labelled objects; submit a per-object probability \cite{demianenko2023understanding}. The work lies in
representing a sparse, gappy, multi-band series rather than in model capacity.
The log-loss component requires calibration, so a good ranking with
overconfident scores is penalised.

\subsection{Neuroscience}

\paragraph{ASD classification (\texttt{abide\_classification}).}
Resting-state fMRI on a $200$-region atlas from sixteen training sites; submit
an autism probability and a chronological age for every subject of four
entirely held-out sites \citep{kan2022brain, popov2024meanmlp}. Scored on
discrimination plus calibration against the training prevalence, with an
auxiliary age term. The ceiling is set by cross-scanner transfer, not by method
quality.

\paragraph{NMM family identification (\texttt{nmm\_forward\_modelling}).}
Forty-five anonymous three-node recordings; name the generating neural-mass
family for each \citep{sanzleon2015virtual, deco2014local}. Equations,
constants and parameter ranges are withheld, only the recorded observable per
family is disclosed, and several series sit in corners of one family's
parameter space that mimic another. The score is chance-corrected, so hedging
earns nothing.

\paragraph{NMM digital twins (\texttt{nmm\_digital\_twins}).}
Spontaneous baseline activity of a 30-region delayed Wilson--Cowan system
\citep{wilsoncowan1972excitatory}, its connectivity and delays, and a future
intervention; submit the signed, time-resolved causal difference between
activity with and without it. Matching resting means, spectra or connectivity
does not determine the response \citep{papadopoulos2020relations}, and the
shared self-excitation gain and per-node drive parameters that set the
operating regime are withheld; the model's remaining coefficients, numerics and
noise scale are published in a manifest.

\paragraph{NMM parameter recovery (\texttt{nmm\_parameter\_recovery}).}
z-scored BOLD per subject; submit global coupling and two receptor-weighted
gain parameters \citep{deco2014local, hansen2022mapping}. The generator is
fully specified in the task description, so the whole difficulty is the inverse
problem: inference over a stochastic delayed mean-field model seen through a
slow, noisy, standardised channel. Standardisation removes amplitude, so only
the regional pattern carries the parameters.

\paragraph{BOLD deconvolution (\texttt{rs\_bold\_deconvolution}).}
Two paired hemodynamic views of one neural realisation, with no stimulus timing
\citep{handwerker2004variation, wu2013blind}; submit the shared latent neural
drive, a double-gamma kernel bank and a Balloon--Windkessel parameter bank
\citep{glover1999deconvolution, friston2000nonlinear}. The two kernel terms
dominate the score and are gated, so a regionally constant template scores
zero. Balloon--Windkessel parameters are credited only up to functional
equivalence, through a frozen ODE on four probes.
\section{Agent-Facing Prompt Surface}
\label{app:prompts}
 
Two texts fully determine what an agent sees before it begins: a system prompt
shared by all twelve tasks, and the task's own \texttt{description.md}. Neither
reveals the grader implementation, the anchors, or any ground truth. We
reproduce the system prompt in full, followed by one complete task prompt. The
remaining eleven are released with the benchmark.
 
\subsection{System prompt}
 
\begin{lstlisting}
You are an autonomous scientific problem-solving agent.
 
Your primary objective is to maximize the scientific score using only the visible workspace data and the tools available in the sandbox. Structural validity and reproducibility are acceptance requirements, not substitutes for solving the scientific problem.
 
Before writing solution files, read the complete task brief and relevant public metadata. Identify the scored scientific target, the variation the metric rewards, and every stated numerical or modelling convention. Inspect the actual data, select a scientifically appropriate method, implement it, and test the scored outputs. A submission can pass structural validation and still score zero. Do not rely on constant, group-midpoint, globally broadcast, or data-ignoring predictions when the metric rewards continuous, sample-specific, regional, or temporal variation.
 
Do not search outside the workspace for existing solutions, hidden answers, grader code, or private data.
 
After inspecting the task and data, create a valid initial submission before starting long training, simulation, or optimization jobs. Keep the required files valid as you improve the scientific method so a budget limit cannot erase all completed work.
 
After that checkpoint is valid, use available budget for substantive scientific improvement. Diagnose the current method from public data, identify its largest unresolved uncertainty, and test a materially different method, modelling choice, or numerical setting. Keep an improvement only when public-data diagnostics support it. Do not spend budget on repeated summaries, unchanged tests, or speculative prose; do not repeat an experiment unless its hypothesis, implementation choice, or diagnostic changes.
 
Once the scientific method produces working scored outputs, create the smallest reproducibility package required by the task. Reuse the tested implementation; do not spend budget polishing documentation or package structure. Produce every scored output and required artifact at its exact relative path, preserving required subdirectories. The final `run.py` must read only visible task data and regenerate the scored outputs in the requested output directory, and `README.md` must give concise usage instructions.
 
Before finishing:
 
1. Run the solution entry point and fix execution errors.
2. Confirm that every required scored output and artifact exists at its exact path.
3. Call `validation_feedback()`, fix every reported structural problem, and validate again.
4. Call `submit()` only after validation succeeds. `submit()` independently rechecks the submission and rejects incomplete or malformed output.
 
If the full method cannot be completed within the budget, submit the strongest scientifically defensible working solution available. Do not replace the scientific method with an empty or deliberately invalid stub.
\end{lstlisting}
 
\subsection{Example task prompt: \texttt{stokes\_inversion}}
 
We reproduce this task because it exercises four of the seven properties of
Section~\ref{sec:unique} at once: an underdetermined solution set with exact
structural non-identifiability (P1), a submission that includes the agent's own
forward model rather than predictions alone (P2), a composite geometrically
gated metric (P5), and a failure mode in which a spectrum is reproduced while
the physics is violated (P6).

\begin{lstlisting}
# Simulated Stokes Inversion

Given a simulated Stokes spectrum on a fixed 56-point wavelength grid around the Fe I 6302.5 Å line, predict the 11 Milne–Eddington atmospheric parameters for each pixel independently, together with an 80% central prediction interval (the 10th / 50th / 90th percentiles).

## Scientific motivation

The Sun's photospheric magnetic field governs every form of solar activity that matters at human timescales — sunspots, the 11-year cycle, flares, coronal mass ejections, and the space-weather drivers that disrupt satellites and power grids. Its only direct remote diagnostic is **spectropolarimetry**: in the presence of a magnetic field, spectral lines split into polarized components through the Zeeman effect, and the four Stokes parameters $(I, Q, U, V)$ encode the full magnetic vector together with the thermodynamic and velocity state of the line-forming layer. Recovering atmospheric parameters from observed Stokes profiles — the **inversion problem** — is therefore the central numerical task of modern solar physics.

In the Milne–Eddington (ME) approximation, the line-forming atmosphere is summarised by 11 parameters: magnetic-field strength $B$, inclination $\theta$, azimuth $\chi$, Doppler width $\Delta_D$, damping $a$, line-strength ratio $\eta_0$, the source-function pair $(S_0, S_1)$, line-of-sight velocity $v_\text{LOS}$, magnetic filling factor $\alpha$, and stray-light velocity $v_\text{stray}$. The forward map from these eleven scalars to the four Stokes profiles is the analytical Unno–Rachkovsky solution of the polarized radiative-transfer equation. Participants are expected to implement their own forward map from the published solar-physics literature (e.g. Landi Degl'Innocenti & Landolfi 2004).

This benchmark isolates this inverse problem in a **fully simulated, fully controlled** setting. The training and test data are produced from the validated ME forward model with channel-dependent Gaussian noise; generating parameters are exactly known to the organizer, and test targets remain private. The challenge is pixelwise: each spectrum is processed independently, with no spatial context. This isolates the per-pixel inversion competence as a clean probe of a method's ability to invert a polarised radiative-transfer model.

### Physical difficulties of ME inversion

Even with perfect training data, the inverse problem has structural difficulties that any method must address:

1. **Non-linearity.** The Unno–Rachkovsky forward map composes Voigt profiles, Faraday–Voigt anomalous-dispersion profiles, and trigonometric functions of the field angles. Small parameter changes can produce discontinuous-looking changes in the spectrum (e.g. when an absorption profile transitions from un-split to split as $B$ crosses a Doppler-width-dependent threshold), and large parameter changes can produce visually identical spectra.
2. **The 180-degree azimuth ambiguity.** The azimuth $\chi$ enters Stokes $Q$ and $U$ only through $\sin(2\chi)$ and $\cos(2\chi)$. The pair $\chi$ and $\chi + 180°$ produce **identical** Stokes profiles, so there is no information in the spectrum that distinguishes them. The benchmark treats $\chi$ as a **circular variable with period 180°** everywhere in scoring (submission validation, pinball loss, interval coverage, interval width).
3. **Structural non-identifiability at singular points.** A few exact generator-defined sentinels make specific parameters genuinely unobservable for a given pixel, as a matter of physics rather than noise:
   - $\alpha = 0$ (no magnetic component): $B, \theta, \chi, v_\text{LOS}$ are unobservable.
   - $B = 0$ (zero field): $\theta, \chi$ are unobservable (no transverse or longitudinal direction).
   - $\alpha = 1$ (no stray component): $v_\text{stray}$ is unobservable.
   These exact-singularity samples are **masked out** of the score for the affected parameters only; near-singular cases remain scored and an honest method can protect itself through appropriately wide conditional intervals.
4. **Polarization signal-to-noise varies with $B$, geometry, $\alpha$, and noise.** Weak-field, near-longitudinal, or low-$\alpha$ pixels have faint polarization and the recovered field direction is noisy — but this is governed by the polarization SNR actually present in the spectrum, not by a fixed $B$ threshold. The dataset contains weak-field pixels with detectable polarization; do not assume a blanket cutoff.
5. **Filling-factor / field-strength degeneracy.** A weak field filling the entire pixel ($\alpha = 1$, small $B$) can produce a circular-polarization amplitude very similar to a strong field filling a fraction of the pixel ($\alpha < 1$, large $B$). When $\alpha \to 1$, $v_\text{stray}$ becomes mathematically undefined.

## Dataset

The published dataset lives under `dataset/`:

| File | Shape | dtype | Role |
|---|---|---|---|
| `spectra_train.npy` | `(100000, 56, 4)` | float32 | absolute-scale Stokes $[I, Q, U, V]$, with noise |
| `params_train.npy`  | `(100000, 11)`   | float64 | ground-truth training parameters |
| `spectra_test.npy`  | `(10000, 56, 4)` | float32 | test spectra (the only test input you receive) |
| `wavelength_grid.npy` | `(56,)` | float64 | mÅ offsets from Fe I 6302.5 Å line center |

`params_test.npy` (shape `(10000, 11)`) is the withheld ground truth used for scoring; it is **not** an input. Load the published arrays directly with `numpy`:

```python
import numpy as np
spectra_train = np.load("spectra_train.npy")   # (100000, 56, 4) float32
params_train  = np.load("params_train.npy")    # (100000, 11)   float64
spectra_test  = np.load("spectra_test.npy")    # (10000, 56, 4) float32
wavelength    = np.load("wavelength_grid.npy") # (56,) mÅ offsets from Fe I 6302.5 Å
```

The harness stages these four arrays plus the `public_manifest.json` dataset manifest — and nothing else — into the agent workspace; the exact mount path is supplied by the STEMBench instruction wrapper. Ground-truth files marked `role: ground_truth` in `config.yaml` (`params_test.npy`, `test_metadata.npz`) are never mounted into the agent sandbox.

### The 11 physical parameters (canonical order — never re-order)

| Idx | Symbol | Name | Unit | Range |
|----:|--------|------|------|-------|
| 0 | `B` | Magnetic field strength | G | $[0, 5000]$ |
| 1 | `theta` | Field inclination from line of sight | deg | $[0, 180]$ |
| 2 | `chi` | Field azimuth in plane of sky (period 180°) | deg | $[0, 180)$ |
| 3 | `Delta_D` | Doppler width | mÅ | $[20, 90]$ |
| 4 | `a` | Damping (Voigt parameter) | — | $[0, 1.5]$ |
| 5 | `eta_0` | Line-strength ratio | — | $[0.01, 100]$ |
| 6 | `S_0` | Source function at $\tau = 0$ | intensity | $[0, 38\,603]$ |
| 7 | `S_1` | Source function gradient | intensity | $[0, 60\,464]$ |
| 8 | `v_LOS` | LOS velocity (magnetic component) | km/s | $[-10, 10]$ |
| 9 | `alpha` | Magnetic filling factor | — | $[0, 1]$ |
| 10 | `v_stray` | LOS velocity (field-free component) | km/s | $[-10, 10]$ |

### Train/test split composition

The split is **globally group-disjoint**: no exact 11-parameter vector and no source row is shared between training, the organizer-only calibration set, and the test set. Both train and test are **balanced across four magnetic-field regimes** of equal test weight:

| Regime | $B$ range (G) |
|---|---|
| weak | $[0, 200)$ |
| intermediate | $[200, 500)$ |
| moderate | $[500, 1500)$ |
| strong | $[1500, 5000]$ |

Each regime contributes an equal share of train and of test, so changes in sample composition cannot silently reweight the scientific question. The four regimes are averaged equally in the score.

The test set additionally contains a **stress** subset (20% of test, 2,000 rows). The stress designation and per-row core/stress labels are **private** (organizer-only); only aggregate core/stress results are reported after evaluation. The stress split preserves the $B$-regime × filling-factor balance and the one-dimensional parameter marginals (audited against the core split) but is a **stratified random diagnostic**, not an interpolation probe: the audit documents that stress joint cells do recur in training within each B regime. The headline score evaluates probabilistic recovery ($R$) and forward consistency ($F$) over the four equally weighted magnetic regimes; core vs stress is reported as a stratified random split diagnostic, not a prediction about which method classes should succeed or fail.

### Spectra, noise, and units

Spectra are stored in **absolute intensity units** (not continuum-normalised), so the continuum level directly anchors the source-function parameters $S_0$ and $S_1$. Channel-dependent additive Gaussian noise is already baked into the released spectra with per-channel sigma

$$\sigma_{I,Q,U,V} = [109,\,2,\,2,\,44]\,/\,20\,000 \;\cdot\; (S_0 + \mu S_1), \qquad \mu = 1,$$

in continuum-intensity units. This is the benchmark noise model — a frozen, reproducible channel-dependent model; it is not an exact Hinode/SOT-SP calibration.

## Submission contract

A submission is a directory containing **three numpy arrays** of shape $(10000, 11)$ **plus a forward-model entry point**:

- `<submission_dir>/pred_q10.npy` — 10th-percentile prediction
- `<submission_dir>/pred_q50.npy` — median (point prediction)
- `<submission_dir>/pred_q90.npy` — 90th-percentile prediction
- `<submission_dir>/forward_model/entry_point.txt` — text file with one line of the form `<module>:<attr>` (e.g. `me:me_model`), naming a callable inside `<submission_dir>/forward_model/`
- `<submission_dir>/forward_model/<module>.py` — your implementation(s) of the callable

The grader **rejects** (scores 0, before any metric is computed) any submission with **structural** violations:

- A missing or extra-shaped array; an object, boolean, integer, NaN, or infinity value.
- Wrong test length or parameter count.

**Parameter-level violations are scored, not rejected** — the pinball metric penalizes them naturally, and the grader report lists them as warnings:

- Any non-circular parameter outside its physical bounds in any of the three quantile files.
- Reversed quantiles: `q10 > q50` or `q50 > q90` for any non-circular entry.
- For `chi`: `q50_chi` outside $[0, 180°)$; `q10_chi > q50_chi` or `q50_chi > q90_chi`; either side extending more than $90°$ from `q50_chi`.

`validation_feedback` reports these parameter issues as warnings so you can fix them before submitting.

### Forward-model contract (for the $F$ term)

The forward-consistency term $F$ synthesizes a noiseless spectrum from your submitted `q50` parameters using **your** forward model — not an organizer-supplied one — and compares it to the observed noisy test spectrum. `forward_model/entry_point.txt` is a **required submission file**; a submission missing it is rejected at validation. To compute $F$, the grader runs your forward code inside an **isolated Docker container** (non-root, read-only filesystem, no network, no access to private ground truth or the task root):

1. Reads `forward_model/entry_point.txt`, parses `<module>:<attr>`.
2. Imports `<module>` from your `forward_model/` directory inside the container.
3. Calls `<attr>(params)` where `params` is `(N, 11) float64` — your `q50`.
4. Expects the return to be `(N, 56, 4) float64` **absolute-scale noiseless** Stokes $[I,Q,U,V]$.

If the callable is present but fails to import, raises, or returns the wrong shape, $F = 0$. $R$ (recovery) is unaffected — you keep the credit for calibrated parameter quantiles even with a non-working forward. Line constants ($\lambda_0 = 6302.5$ Å, $g_\text{eff} = 2.5$, $\mu = 1$; the 56-point wavelength grid in mÅ spans 6302.069–6303.254 Å) must be embedded in your forward model code — either hard-coded or imported from a constants module shipped inside your `forward_model/` directory. At grade time the grader's own `grader/constants.py` is available on the import path inside the synthesis container; the canonical values are `LINE_VEC = (6302.5, 2.5, 1)` and `LINE_ARG = 1000 * (np.linspace(6302.0692255, 6303.2544205, 56) - 6302.5)` (in mÅ).

Provide a physical synthesis model that preserves the meanings and units of the 11 inputs. The data were generated with a Milne–Eddington / Unno–Rachkovsky atmosphere, but your implementation may use a richer, stratified or multicomponent atmosphere. Supply any additional atmospheric quantities deterministically within your forward model and document their relationship to the 11 inputs. The grader does not compare your implementation with a prescribed solver.

Submitted forward models also pass physical-validity probes (the $P_\text{phys}$ factor of the score); their specification is in `metric_description.md`. They check the physical meanings of the declared interface — not agreement with any particular ME solver.

### Chi (circular) interval convention

$\chi$ keeps the three-file contract but its endpoints live on a local unwrapped coordinate:

- `q50_chi` is the circular center, in $[0, 180°)$.
- `q10_chi ≤ q50_chi ≤ q90_chi`, and each side may extend at most $90°$ from `q50_chi`.
- `q10_chi` and `q90_chi` **may fall outside** $[0, 180°)$ so an interval can cross the $0/180°$ boundary.

During scoring, the true $\chi$ is lifted to the equivalent value nearest `q50` before the pinball loss, interval coverage, and interval width are computed. This makes every $\chi$ operation circular while preserving the three-file submission format. Predictions that differ only by a $180°$ shift in $\chi$ score identically.

If you only produce a point estimate, set `q10 = q50 = q90` (a zero-width interval). This is valid but earns an uncertainty penalty under the pinball metric — calibrated intervals score better.

## Allowed Tools and Libraries

The sandbox has no network access, so only the packages pre-installed in the sandbox image can be used. Packages relevant to this task:

- `numpy` — numerical arrays and core math for the spectral cubes.
- `scipy` — scientific routines (optimization, interpolation, statistics).
- `scikit-learn` — classical machine-learning models and preprocessing.
- `xgboost` — gradient-boosted decision trees.
- `catboost` — gradient boosting on tabular features.
- `torch` — neural network models.
- `numba` — JIT compilation for fast pixelwise numerical loops.

## Scoring

A submission is scored on the test split as a single scalar in $[0, 1]$, **higher is better**, perfect = 1:

$$\text{FinalScore} \;=\; 0.80 \cdot R \;+\; 0.20 \cdot \sqrt{R \cdot F_\text{eff}}$$

where $R$ is a baseline-normalized, null-gated **probabilistic recovery** score and $F$ is a **forward-consistency** score. A constant train-marginal predictor scores at the floor (its raw skill is ~0 by construction and the null gate floors the residual); an oracle scores $1$.

### $R$ — probabilistic recovery (primary)

For each magnetic-field regime $r$ and parameter $k$, the grader computes the mean pinball loss across $\tau \in \{0.1, 0.5, 0.9\}$ over the observable sample–parameter pairs (exact singularity pairs are masked; $\chi$ is handled circularly):

$$L_{r,k} = \frac{1}{3}\sum_{\tau \in \{0.1, 0.5, 0.9\}} \overline{\rho_\tau(y_{i,k},\, \hat q^{(\tau)}_{i,k})}, \qquad
\rho_\tau(y, \hat q) = \max\!\bigl(\tau(y-\hat q),\, (\tau-1)(y-\hat q)\bigr).$$

Let $L_{0,r,k}$ be the pinball loss of the constant train-marginal quantile predictor (the no-skill baseline) on the test split. The raw skill is baseline-normalized:

$$S^\text{raw}_{r,k} = \mathrm{clip}\!\left(1 - L_{r,k} / L_{0,r,k},\ 0,\ 1\right),$$

then null-gated and rescaled above a frozen null threshold $t_{r,k}$ (the 95th percentile of raw skill across $\geq 1000$ null submissions: label permutations, shuffled good predictions, random bounded predictions):

$$S_{r,k} = \begin{cases} 0 & S^\text{raw}_{r,k} \le t_{r,k} \\ (S^\text{raw}_{r,k} - t_{r,k})/(1 - t_{r,k}) & \text{otherwise.}\end{cases}$$

Within each regime, skills are grouped by physical role and weighted:

$$S_\text{mag} = \mathrm{mean}(S_B, S_\theta, S_\chi, S_\alpha),\quad
S_\text{thermo} = \mathrm{mean}(S_{\Delta_D}, S_a, S_{\eta_0}, S_{S_0}, S_{S_1}),\quad
S_\text{vel} = \mathrm{mean}(S_{v_\text{LOS}}, S_{v_\text{stray}}),$$
$$R_r = 0.50\,S_\text{mag} + 0.30\,S_\text{thermo} + 0.20\,S_\text{vel}, \qquad
R = \mathrm{mean}(R_\text{weak}, R_\text{intermediate}, R_\text{moderate}, R_\text{strong}).$$

The grader applies frozen monotone transforms internally before computing pinball loss — $\log1p$ on $B$ and $\eta_0$, identity on the rest — so quantiles are equivariant and the score is invariant to test-composition changes. Pearson correlation is **not** a component of the score.

### $F$ — forward consistency (physical-equivalence gate)

For each test row, the grader synthesizes a noiseless spectrum from the submitted `q50` parameters using **the participant's submitted forward model** (declared in `forward_model/entry_point.txt`), and compares it to the observed noisy test spectrum with the per-sample channel noise:

$$D_i = \mathrm{mean}_{\lambda, c}\ \mathrm{Huber}\!\left(\frac{x_\text{obs} - x_\text{pred}(q_{50})}{\sigma},\ \delta\!=\!3\right).$$

Per regime, $F_r = \mathrm{clip}\!\left((D_0 - D_r)/(D_0 - D_\text{oracle}),\ 0, 1\right)$, where $D_0$ is the constant-baseline residual and $D_\text{oracle}$ the hidden-truth residual; then $F = \mathrm{mean}_r F_r$. The task is to provide a genuine physical synthesis model, and forward consistency is scaled by physical validity $P_\text{phys} \in [0, 1]$ ($F_\text{eff} = F \cdot P_\text{phys}$). $F$ rewards a physically equivalent inversion (good spectral fit) but the geometric gate $\sqrt{R \cdot F_\text{eff}}$ ensures a pure forward fit with arbitrary latent parameters cannot win on its own — $F$ only helps when recovery is already non-trivial.

### Reported diagnostics (not additional score terms)

The grader report includes, beyond `FinalScore`: $R$ and $F$; per-parameter pinball skill; the four $B$-regime scores; core vs stress scores; $q_{10}$–$q_{90}$ coverage and width; circular-$\chi$ coverage and width; median MAE in physical units; polarization-SNR strata; exact-singularity counts and masks; forward residual by Stokes channel.

## Evaluation

The STEMBench harness scores a submission automatically by calling
`grade(submission_dir, answers_path)` from `grade.py`. No scoring command is
issued by the participant. To produce a scoreable submission, write exactly
three files to your solution directory:

- `pred_q10.npy` — shape `(10000, 11)`, float32/float64
- `pred_q50.npy` — shape `(10000, 11)`, float32/float64
- `pred_q90.npy` — shape `(10000, 11)`, float32/float64

Ground truth is never mounted into the agent sandbox; the harness reads it
host-side.
\end{lstlisting}
 
 
\section{Run Configuration}
\label{app:runconfig}

Table~\ref{tab:runconfig} specifies the per-episode configuration used for
every run reported in Sections~\ref{sec:results} and~\ref{sec:effort},
unless explicitly stated otherwise (the extended-limits ablation of
Table~\ref{tab:budget} is the only exception). All fields are harness-level
run policy, declared at launch and logged in every episode's metadata; none
of them is part of a task's \texttt{config.yaml}; the wall-clock, token and message budgets are restated to the agent in its instructions.

\begin{table}[h]
\caption{Per-episode run configuration. ``Value'' is the setting used for
all reported runs; ``Semantics'' states what the field controls and how the
harness enforces it.}
\label{tab:runconfig}
\centering
\small
\setlength{\tabcolsep}{4pt}
\begin{tabular}{@{}l l p{8.3cm}@{}}
\toprule
\textbf{Field} & \textbf{Value} & \textbf{Semantics} \\
\midrule
\texttt{sandbox} & \texttt{docker-cuda} &
Offline per-domain container (network disabled, public data mounted
read-only, writable solution directory).  \\
\texttt{time\_limit\_sec} & $3\,600$ &
Wall-clock budget per episode. The agent loop is cancelled at the deadline
and whatever submission is present at that moment is graded; a single
\texttt{bash}/\texttt{python} execution may run for at most two thirds of
the budget. \\
\texttt{token\_limit} & $10^6$ &
Cumulative model tokens (input plus output) across all model calls of the
episode; the next model call is refused once the budget is spent. \\
\texttt{max\_messages} & $200$ &
Total conversation messages (user, assistant and tool results) per episode;
the episode ends when the limit is reached. \\
\texttt{max\_tool\_calls} & $200$ &
Declared tool-call budget, recorded in the run plan and episode metadata.
It is not an independent stop condition: every tool call adds messages, so
the message limit binds first. \\
\texttt{effort} & low / medium / high &
Reasoning-effort tier forwarded to the model API as
\texttt{reasoning.effort} through the OpenRouter gateway; the experimental
variable of Section~\ref{sec:effort}, with \texttt{high} the harness
default. \\
\bottomrule
\end{tabular}
\end{table}

All campaign episodes executed on single-GPU instances:
one Tesla~T4 (16~GB) on an Intel Ice~Lake platform with 8~vCPU and
32~GB RAM, Ubuntu~24.04, with the per-domain CUDA container image holding
the only software the agent can touch. 

Model access uses an OpenAI-compatible protocol through the OpenRouter
gateway; sampling parameters (temperature, top-$p$) are not exposed on this
route for reasoning models and were not set, which is why run-to-run
variability is measured by repetition.

\paragraph{Harness and contract tests.}
The harness is covered by a functional test suite over the tool layer,
which passes $17/17$ checks: the six-tool registry, the eight file formats
that \texttt{data\_explorer} advertises, the refusal of unsafe
deserialisation, valid and invalid submission checks (including
split-keyed JSON and per-key \texttt{.npz} constraints), and the mapping
from the submission checks in \texttt{config.yaml} to validator
specifications. All twelve task contracts additionally pass
dataset-checksum verification and smoke tests that grade the shipped
baseline and expert submissions and reproduce the anchors in
Table~\ref{tab:tasks}.

\section{Per-Model Results}
\label{app:results}

Table~\ref{tab:results} reports eight frontier models under the \textbf{basic}
scaffold at high reasoning effort. Scores are the task's native headline
metric in $[0,1]$ under the current task revisions, and each cell is the mean over three retained scored runs, including zeros.
 
\begin{table}[h]
\caption{\bench{} results, \textbf{basic} agent, raw task scores.
Each cell is the mean over three scored high-effort runs retained
in the current report (288 runs in total), including zeros.
Domain means are unweighted means of the task means, computed before rounding.
Bold denotes the highest unrounded mean in each row.
$b$/$e$ are the baseline and expert anchors.
$^{\dagger}$The digital-twin expert anchor is $e=0.715233$
(displayed as $0.72$).}
\label{tab:results}
\centering\small
\setlength{\tabcolsep}{3pt}
\resizebox{\textwidth}{!}{%
\begin{tabular}{@{}l cc cccccccc@{}}
\toprule
\textbf{Task} & \textbf{$b$} & \textbf{$e$}
& \textbf{Gemini 3.7 Flash} & \textbf{Gemini 3.1 Pro}
& \textbf{Opus 5} & \textbf{DeepSeek V4 Pro}
& \textbf{Qwen3.8 2.4T} & \textbf{GLM 5.2}
& \textbf{MiMo V2.5} & \textbf{MiniMax M3} \\
\midrule
New-physics search & 0.33 & 0.91 & 0.31 & \textbf{0.93} & 0.53 & 0.83 & 0.54 & 0.62 & 0.00 & 0.38 \\
Higgs uncertainty & 0.40 & 0.69 & 0.41 & 0.43 & \textbf{0.48} & 0.39 & 0.44 & 0.28 & 0.14 & 0.13 \\
Top tagging & 0.06 & 0.88 & 0.56 & 0.65 & \textbf{0.94} & 0.89 & 0.93 & 0.60 & 0.59 & 0.56 \\
Dark Machines & 0.00 & 0.22 & 0.22 & 0.29 & 0.25 & 0.28 & 0.22 & \textbf{0.32} & 0.18 & 0.20 \\
\textit{HEP mean} & --- & --- & 0.37 & 0.57 & 0.55 & \textbf{0.60} & 0.53 & 0.46 & 0.23 & 0.32 \\
\midrule
Stokes, simulated & 0.01 & 0.62 & 0.01 & \textbf{0.78} & 0.57 & 0.25 & 0.00 & 0.00 & 0.43 & 0.11 \\
Stokes, real & 0.00 & 0.60 & 0.12 & \textbf{0.82} & 0.09 & 0.00 & 0.00 & 0.00 & 0.14 & 0.00 \\
SN~Ia classification & 0.42 & 0.85 & 0.50 & 0.91 & \textbf{0.94} & 0.90 & 0.93 & 0.60 & 0.93 & 0.93 \\
\textit{Astro mean} & --- & --- & 0.21 & \textbf{0.84} & 0.53 & 0.38 & 0.31 & 0.20 & 0.50 & 0.34 \\
\midrule
ASD classification & 0.00 & 0.31 & 0.09 & 0.00 & 0.00 & 0.10 & 0.10 & \textbf{0.29} & 0.08 & 0.19 \\
NMM family id. & 0.00 & 0.90 & 0.00 & \textbf{1.00} & 0.00 & 0.51 & 0.56 & 0.00 & 0.37 & 0.02 \\
NMM digital twins$^{\dagger}$ & 0.10 & 0.72 & 0.10 & 0.24 & \textbf{0.31} & 0.14 & 0.10 & 0.03 & 0.00 & 0.10 \\
NMM param.\ recovery & 0.12 & 0.51 & 0.12 & 0.18 & \textbf{0.42} & 0.12 & 0.12 & 0.08 & 0.02 & 0.12 \\
BOLD deconvolution & 0.01 & 0.35 & 0.00 & \textbf{0.19} & 0.00 & 0.00 & 0.02 & 0.00 & 0.00 & 0.06 \\
\textit{Neuro mean} & --- & --- & 0.06 & \textbf{0.32} & 0.15 & 0.18 & 0.18 & 0.08 & 0.09 & 0.10 \\
\bottomrule
\end{tabular}}
\end{table}

Table~\ref{tab:results-max} complements Table~\ref{tab:results} with the
\emph{maximum} of the same three retained runs per cell. The two views
separate ability from reliability: on the HEP tasks six of the eight models
reach a best-run mean of $0.63$--$0.71$, so the spread of the means in
Table~\ref{tab:results} is driven largely by how often a model fails, not by
whether it can solve the task at all. The individual-run records quoted in
Section~\ref{sec:results} appear here as cell maxima: ABIDE at $0.31$
($0.314$ unrounded, DeepSeek~V4~Pro) and parameter recovery at $0.58$
($0.583$, Opus~5).

\begin{table}[h]
\caption{\bench{} results, \textbf{basic} agent, \emph{maximum} raw task
scores. Each cell is the maximum over the same three retained high-effort
runs as in Table~\ref{tab:results}, including zeros. Domain rows are unweighted means
of the task maxima, computed before rounding. Bold denotes the highest
unrounded maximum in each row. $b$/$e$ are the baseline and expert anchors.
$^{\dagger}$The digital-twin expert anchor is $e=0.715233$
(displayed as $0.72$).}
\label{tab:results-max}
\centering\small
\setlength{\tabcolsep}{3pt}
\resizebox{\textwidth}{!}{%
\begin{tabular}{@{}l cc cccccccc@{}}
\toprule
\textbf{Task} & \textbf{$b$} & \textbf{$e$}
& \textbf{Gemini 3.7 Flash} & \textbf{Gemini 3.1 Pro}
& \textbf{Opus 5} & \textbf{DeepSeek V4 Pro}
& \textbf{Qwen3.8 2.4T} & \textbf{GLM 5.2}
& \textbf{MiMo V2.5} & \textbf{MiniMax M3} \\
\midrule
New-physics search & 0.33 & 0.91 & 0.50 & \textbf{0.94} & 0.93 & 0.94 & 0.82 & 0.94 & 0.00 & 0.94 \\
Higgs uncertainty & 0.40 & 0.69 & 0.41 & 0.49 & \textbf{0.53} & 0.47 & 0.47 & 0.44 & 0.41 & 0.38 \\
Top tagging & 0.06 & 0.88 & 0.80 & 0.95 & \textbf{0.95} & 0.93 & 0.94 & 0.93 & 0.91 & 0.87 \\
Dark Machines & 0.00 & 0.22 & 0.33 & 0.33 & 0.38 & \textbf{0.48} & 0.35 & 0.34 & 0.32 & 0.32 \\
\textit{HEP mean} & --- & --- & 0.51 & 0.68 & 0.70 & \textbf{0.71} & 0.64 & 0.66 & 0.41 & 0.63 \\
\midrule
Stokes, simulated & 0.01 & 0.62 & 0.01 & 0.80 & \textbf{0.85} & 0.74 & 0.01 & 0.00 & 0.74 & 0.32 \\
Stokes, real & 0.00 & 0.60 & 0.18 & \textbf{0.86} & 0.15 & 0.00 & 0.00 & 0.00 & 0.41 & 0.00 \\
SN~Ia classification & 0.42 & 0.85 & 0.50 & 0.93 & \textbf{0.95} & 0.93 & 0.94 & 0.95 & 0.94 & 0.93 \\
\textit{Astro mean} & --- & --- & 0.23 & \textbf{0.87} & 0.65 & 0.56 & 0.31 & 0.32 & 0.70 & 0.42 \\
\midrule
ASD classification & 0.00 & 0.31 & 0.28 & 0.00 & 0.00 & \textbf{0.31} & 0.31 & 0.31 & 0.25 & 0.24 \\
NMM family id. & 0.00 & 0.90 & 0.00 & \textbf{1.00} & 0.00 & 0.87 & 0.87 & 0.00 & 0.87 & 0.07 \\
NMM digital twins$^{\dagger}$ & 0.10 & 0.72 & 0.10 & 0.26 & \textbf{0.56} & 0.23 & 0.10 & 0.10 & 0.00 & 0.10 \\
NMM param.\ recovery & 0.12 & 0.51 & 0.12 & 0.30 & \textbf{0.58} & 0.12 & 0.12 & 0.12 & 0.07 & 0.12 \\
BOLD deconvolution & 0.01 & 0.35 & 0.00 & \textbf{0.27} & 0.00 & 0.00 & 0.07 & 0.00 & 0.00 & 0.18 \\
\textit{Neuro mean} & --- & --- & 0.10 & \textbf{0.37} & 0.23 & 0.31 & 0.30 & 0.11 & 0.24 & 0.14 \\
\bottomrule
\end{tabular}}
\end{table}


\section{Pilot Reasoning-Effort Sweep}
\label{sec:effort}
 
This appendix reports a pilot sweep run before the cross-model campaign of
Section~\ref{sec:results}. Its absolute scores are not comparable with that
campaign for two reasons: several task contracts were revised afterwards,
and the runs predate a fix to GPU access in the task containers. We report
it for what it shows about run-to-run variance and the non-monotone effect
of reasoning effort within one model, not as a measurement of the current
benchmark. After the pilot, all campaign runs were fixed at \texttt{high}
effort.

We evaluate \texttt{google/gemini-3.1-pro-preview} with the \texttt{basic}
scaffold and the limits of Appendix~\ref{app:runconfig}, varying only the \emph{reasoning effort}
(\texttt{low}, \texttt{medium}, \texttt{high}): 12 tasks $\times$ 3 levels
$\times$ 3 runs, with the \texttt{top\_quark\_tagging} high-effort cell
extended by 11 runs to estimate single-run variance, for 119 graded runs.
Runs interrupted by infrastructure (proxy) outages were flagged automatically
and excluded; zeros in retained runs are genuine episode outcomes.
 
\begin{table}[h]
\caption{Agent scores by task and reasoning effort, \texttt{google/gemini-3.1-pro-preview}
on the \texttt{basic} scaffold. Cells give mean $\pm$ standard deviation over three runs,
except \texttt{top\_quark\_tagging} at high effort ($n=14$, marked $^{\dagger}$).
``Best'' is the maximum over all runs of the task; ``$0$-runs'' counts runs scoring exactly zero: a missing submission, a
submission that failed validation, or a valid submission that did not
outperform the no-skill baseline; the count does not distinguish these
cases.
The last rows aggregate the twelve per-task means.}
\label{tab:agg}
\centering
\small
\setlength{\tabcolsep}{5pt}
\begin{tabular}{@{}l l ccc cc@{}}
\toprule
& \textbf{Task} & \textbf{low} & \textbf{medium} & \textbf{high} &
\textbf{Best} & \textbf{$0$-runs} \\
\midrule
\multirow{4}{*}{HEP} & New-physics search & $0.3 \pm 0.3$ & $0.3 \pm 0.5$ & $0.6 \pm 0.5$ & $0.932$ & $4/9$ \\
 & Higgs uncertainty & $0.44 \pm 0.05$ & $0.46 \pm 0.16$ & $0.5 \pm 0.1$ & $0.640$ & $0/9$ \\
 & Top tagging & $0.806 \pm 0.006$ & $0.86 \pm 0.03$ & $0.8 \pm 0.2^{\dagger}$ & $0.941$ & $0/20$ \\
 & Dark Machines & $0.22 \pm 0.19$ & $0.31 \pm 0.03$ & $0.32 \pm 0.05$ & $0.359$ & $0/9$ \\
\midrule
\multirow{3}{*}{Astro} & SN~Ia classification & $0.73 \pm 0.02$ & $0.87 \pm 0.03$ & $0.91 \pm 0.04$ & $0.935$ & $0/9$ \\
 & Stokes, simulated & $0.07 \pm 0.11$ & $0.74 \pm 0.07$ & $0.5 \pm 0.4$ & $0.815$ & $0/9$ \\
 & Stokes, real & $0.000 \pm 0.000$ & $0.3 \pm 0.3$ & $0.81 \pm 0.03$ & $0.831$ & $4/9$ \\
\midrule
\multirow{5}{*}{Neuro} & NMM family id. & $0.000 \pm 0.000$ & $1.000 \pm 0.000$ & $1.000 \pm 0.000$ & $1.000$ & $3/9$ \\
 & NMM digital twins & $0.27 \pm 0.15$ & $0.16 \pm 0.09$ & $0.18 \pm 0.05$ & $0.381$ & $0/9$ \\
 & NMM param.\ recovery & $0.125 \pm 0.000$ & $0.2 \pm 0.3$ & $0.10 \pm 0.04$ & $0.468$ & $2/9$ \\
 & BOLD deconvolution & $0.16 \pm 0.14$ & $0.28 \pm 0.13$ & $0.06 \pm 0.09$ & $0.410$ & $1/9$ \\
 & ASD classification & $0.13 \pm 0.16$ & $0.327 \pm 0.000$ & $0.000 \pm 0.000$ & $0.327$ & $4/9$ \\
\midrule
\multicolumn{2}{@{}l}{\textit{Mean of task means}} & $\mathbf{0.272}$ & $\mathbf{0.484}$ & $\mathbf{0.484}$ & & \\
\multicolumn{2}{@{}l}{\textit{Median of task means}} & $0.190$ & $0.333$ & $0.499$ & & \\
\bottomrule
\end{tabular}
\end{table}
 
\begin{table}[H]
\caption{Budget ablation on \texttt{top\_quark\_tagging}. ``Main'' is the primary grid
(1\,h, $10^6$ tokens, 200 messages, 200 tool calls); ``Extended'' raises these to 2.5\,h,
$2\times10^6$ tokens and 500/500, executed through Inspect's native OpenRouter provider.
Mean $\pm$ standard deviation, with the observed range in brackets; $n=3$ per
cell. The fourteen-run variance study of the Main/high cell is reported in
Table~\ref{tab:agg}.}
\label{tab:budget}
\centering
\small
\begin{tabular}{@{}l ll@{}}
\toprule
\textbf{Effort} & \textbf{Main} & \textbf{Extended} \\
\midrule
low & $0.806 \pm 0.006$ $[0.802, 0.813]$ & $0.802 \pm 0.000$ $[0.801, 0.802]$ \\
medium & $0.86 \pm 0.03$ $[0.829, 0.892]$ & $0.85 \pm 0.05$ $[0.802, 0.901]$ \\
high & $0.84 \pm 0.16$ $[0.652, 0.931]$ & $0.941 \pm 0.004$ $[0.937, 0.946]$ \\
\bottomrule
\end{tabular}
\end{table}
 
Table~\ref{tab:agg} aggregates the grid. Three patterns are visible. First,
reasoning effort buys most of its improvement between low and medium
(mean of task means $0.272 \to 0.484$) and nothing on average beyond it
($0.484$ at high), although the median continues to rise
($0.190 \to 0.333 \to 0.499$), so high effort helps the typical task while
losing ground on a few. Second, the dispersion is not noise around a single
mode: several tasks are bimodal, with runs landing either near the agent's
best score or exactly at zero, which is what drives the large standard
deviations on \texttt{lhco\_anomaly} ($\pm 0.53$ at both medium and high) and
\texttt{stokes\_inversion} ($\pm 0.44$ at high). Four of nine runs on each of
\texttt{lhco\_anomaly}, \texttt{hinode\_inversion} and
\texttt{abide\_classification} produced nothing that cleared validation
or the no-skill baseline. Third, the effort--score relation is not monotone per task:
\texttt{abide\_classification} scores $0.327 \pm 0.000$ at medium and
$0.000$ in all three high-effort runs, and \texttt{rs\_bold\_deconvolution}
degrades from $0.276$ at medium to $0.055$ at high. These aggregate scores do not establish whether the declines arise
from invalid submissions, valid low-scoring predictions, or truncation;
distinguishing these explanations requires run traces and validity flags.
 
Table~\ref{tab:budget} compares the main and extended-budget runs. Under extended budgets all three high-effort runs on
\texttt{top\_quark\_tagging} land in $[0.937, 0.946]$
(mean $0.941 \pm 0.004$), above the main-budget mean, and the lower mode
visible in the fourteen-run variance study of Table~\ref{tab:agg}
(runs at $0.80$ and $0.063$) does not appear. At low effort the extended limits change nothing ($0.802 \pm 0.000$), since
low-effort agents stop early on their own. This is consistent with truncation
contributing to high-effort variance, but the extended series also switched
to Inspect's native OpenRouter provider, and the missing GPU acceleration
may itself have caused part of the truncation; the ablation therefore does
not isolate the effect of the budget.

\end{document}

%% file: math_commands.tex
\usepackage{amsmath,amsfonts,bm}

\def\eqref#1{equation~\ref{#1}}

\def\1{\bm{1}}

\DeclareMathAlphabet{\mathsfit}{\encodingdefault}{\sfdefault}{m}{sl}
\SetMathAlphabet{\mathsfit}{bold}{\encodingdefault}{\sfdefault}{bx}{n}



%% file: iclr2025_conference.bib
@article{miller2025biomlbench,
  title = {{BioML-bench}: Evaluation of {AI} Agents for End-to-End Biomedical {ML}},
  author = {Miller, Henry E. and Greenig, Matthew and Tenmann, Benjamin and Wang, Bo},
  journal = {bioRxiv},
  year = {2025},
  doi = {10.1101/2025.09.01.673319},
  url = {https://doi.org/10.1101/2025.09.01.673319}
}

@misc{qiu2026prbench,
  title = {{PRBench}: End-to-end Paper Reproduction in Physics Research},
  author = {Qiu, Shi and Deng, Junyi and Deng, Yiwei and Dong, Haoran and others},
  year = {2026},
  eprint = {2603.27646},
  archivePrefix = {arXiv},
  url = {https://arxiv.org/abs/2603.27646}
}

@misc{chen2025mlrbench,
  title = {{MLR-Bench}: Evaluating {AI} Agents on Open-Ended Machine Learning Research},
  author = {Chen, Hui and Xiong, Miao and Lu, Yujie and Han, Wei and Deng, Ailin and He, Yufei and Wu, Jiaying and Li, Yibo and Liu, Yue and Hooi, Bryan},
  year = {2025},
  eprint = {2505.19955},
  archivePrefix = {arXiv},
  url = {https://arxiv.org/abs/2505.19955}
}

@misc{waters2026composite,
  title = {{COMPOSITE-Stem}},
  author = {Waters, Kyle and Nuzzi, Lucas and Looram, Tadhg and Tomasiello, Alessandro and Kemogne Kamdoum, Ariel Ghislain and Li, Bikun and Sileo, Damien and Kretov, Egor and Fournier-Facio, Francesco and Soloupis, Georgios and Kassahun, Haile and Wolff, Hew and Cai, Jiaqi and Li, Lianghui and Roth, Marc and Naiya, Mohinder and Guo, Naixu and Tang, Qicheng and Wheeler, Richard and Sala, Samuele and Popov, Serguei and Dillmann, Steven and Li, Yuqi},
  year = {2026},
  eprint = {2604.09836},
  archivePrefix = {arXiv},
  url = {https://arxiv.org/abs/2604.09836}
}

@techreport{liu2026lifescibench,
  title = {{LifeSciBench}: Evaluating Language Models on Realistic, Expert-Level Tasks in the Life Sciences},
  author = {Liu, Amelia and Ho, Andrew and Droste, Anne Marie and Martin, David and Wong, Edmund and Zhou, Edward and Zhou, Isabelle and Park, Joshua and Jiao, Joy and Skelly, Katie-Rose and Kim, Kenny and Rao, Kevin and Uehara, Masatoshi and Marion, Max and Fitzgerald, Nicole and Dias, Rachel and Shringarpure, Suyash and Yuan, Yuan and Wang, Yunyun},
  institution = {OpenAI and Tacit Labs},
  year = {2026},
  url = {https://cdn.openai.com/pdf/b4299379-0a97-4ffa-8b9b-c3fbb299caa9/lifescibench_preprint.pdf}
}

@dataset{anonymous_2026_22960135,
  author       = {Anonymous},
  title        = {STEMBench},
  month        = sep,
  year         = 2026,
  publisher    = {Zenodo},
  doi          = {10.5281/zenodo.22960135},
  url          = {https://doi.org/10.5281/zenodo.22960135},
}

@article{zou2026labosbench,
  title={LabOSBench: Benchmarking Computer Use Agents for Scientific Instrument Control},
  author={Zou, Anqi and Deng, Han and Zhang, Chengyu and Hu, Junquan and Wang, Yu and Xing, Yuxiang and Zhang, Aokai and Zhang, Hanling and Liu, Zhaoyang and Fei, Ben and others},
  journal={arXiv preprint arXiv:2606.16802},
  year={2026}
}

@inproceedings{sun2026scienceboard,
  title={Scienceboard: Evaluating multimodal autonomous agents in realistic scientific workflows},
  author={Sun, Qiushi and Liu, Zhoumianze and Ma, Chang and Ding, Zichen and Xu, Fangzhi and Yin, Zhangyue and Zhao, Haiteng and Wu, Zhenyu and Cheng, Kanzhi and Liu, Zhaoyang and others},
  booktitle={International Conference on Learning Representations},
  volume={2026},
  pages={75694--75731},
  year={2026}
}

@software{anonymous_2026_22959834,
  author       = {Anonymous},
  title        = {STEMBench: an Inspect-AI benchmark harness for
                   research-grade STEM agent tasks
                  },
  month        = sep,
  year         = 2026,
  publisher    = {Zenodo},
  version      = {v1.0.0},
  doi          = {10.5281/zenodo.22959834},
  url          = {https://doi.org/10.5281/zenodo.22959834},
  swhid        = {swh:1:dir:855e90123497b68999d647416c3e3e46b607c360
                   ;origin=https://doi.org/10.5281/zenodo.22959833;vi
                   sit=swh:1:snp:770b970b5022d886b877da41fd1cf75fd50a
                   842c;anchor=swh:1:rel:991f4712397ec803272dea7bdec7
                   8b4dfd7a784c;path=/
                  },
}

@article{rein2023gpqa,
  title={Gpqa: A graduate-level google-proof q\&a benchmark},
  author={Rein, David and Hou, Betty Li and Stickland, Asa Cooper and Petty, Jackson and Pang, Richard Yuanzhe and Dirani, Julien and Michael, Julian and Bowman, Samuel R},
  journal={arXiv preprint arXiv:2311.12022},
  year={2023}
}

@article{qiu2025phybench,
  title={Phybench: Holistic evaluation of physical perception and reasoning in large language models},
  author={Qiu, Shi and Guo, Shaoyang and Song, Zhuo-Yang and Sun, Yunbo and Cai, Zeyu and Wei, Jiashen and Luo, Tianyu and Yin, Yixuan and Zhang, Haoxu and Hu, Yi and others},
  journal={arXiv preprint arXiv:2504.16074},
  year={2025}
}

@article{wang2026naturebench,
  title={NatureBench: Can Coding Agents Match the Published SOTA of Nature-Family Papers?},
  author={Wang, Yuru and Cheng, Lejun and Zuo, Yuxin and Zeng, Sihang and He, Bingxiang and Jiang, Che and Yang, Junlin and Wang, Yuchong and Zhao, Kaikai and Huang, Weifeng and others},
  journal={arXiv preprint arXiv:2606.24530},
  year={2026}
}

@article{xu2025ugphysics,
  title={Ugphysics: A comprehensive benchmark for undergraduate physics reasoning with large language models},
  author={Xu, Xin and Xu, Qiyun and Xiao, Tong and Chen, Tianhao and Yan, Yuchen and Zhang, Jiaxin and Diao, Shizhe and Yang, Can and Wang, Yang},
  journal={arXiv preprint arXiv:2502.00334},
  year={2025}
}

@inproceedings{wu2026innovatorbench,
  title={Innovatorbench: evaluating agents’ ability to conduct innovative AI research},
  author={Wu, Yunze and Fu, Dayuan and Si, Weiye and Huang, Zhen and Jiang, Mohan and Li, Keyu and Xia, Shijie and Sun, Jie and Xu, Tianze and Xiao, Yang and others},
  booktitle={International Conference on Learning Representations},
  volume={2026},
  pages={36144--36185},
  year={2026}
}

@software{Terminal-Bench-Science_2026,
  author = {{Terminal-Bench-Science Team}},
  doi = {10.5281/zenodo.22110253},
  license = {Apache-2.0},
  title = {{Terminal-Bench-Science: Evaluating AI agents on research workflows across scientific domains}},
  url = {https://github.com/harbor-framework/terminal-bench-science},
  year = {2026}
}

@article{siegel2024core,
  title={Core-bench: Fostering the credibility of published research through a computational reproducibility agent benchmark},
  author={Siegel, Zachary S and Kapoor, Sayash and Nadgir, Nitya and Stroebl, Benedikt and Narayanan, Arvind},
  journal={arXiv preprint arXiv:2409.11363},
  year={2024}
}

@inproceedings{chen2025scienceagentbench,
  title={Scienceagentbench: Toward rigorous assessment of language agents for data-driven scientific discovery},
  author={Chen, Ziru and Chen, Shijie and Ning, Yuting and Zhang, Qianheng and Wang, Boshi and Yu, Botao and Li, Yifei and Liao, Zeyi and Wei, Chen and Lu, Zitong and others},
  booktitle={International Conference on Learning Representations},
  volume={2025},
  pages={96934--96990},
  year={2025}
}

@article{trofimova2026ml2b,
  title={ML2B: Benchmarking LLMs on Cross-Lingual ML Pipeline Generation},
  author={Trofimova, Ekaterina and Shamina, Zosya and Selifanova, Maria and Zaitsev, Artem and Savchuk, Remi and Minets, Maxim and Ozerova, Daria and Sataev, Emil and Zuenko, Denis and Ustyuzhanin, Andrey},
  year={2026}
}

@inproceedings{chan2025mle,
  title={Mle-bench: Evaluating machine learning agents on machine learning engineering},
  author={Chan, Jun Shern and Chowdhury, Neil and Jaffe, Oliver and Aung, James and Sherburn, Dane and Mays, Evan and Starace, Giulio and Liu, Kevin and Maksin, Leon and Patwardhan, Tejal and others},
  booktitle={International Conference on Learning Representations},
  volume={2025},
  pages={50466--50494},
  year={2025}
}

@inproceedings{
starace2025paperbench,
title={PaperBench: Evaluating {AI}{\textquoteright}s Ability to Replicate {AI} Research},
author={Giulio Starace and Oliver Jaffe and Dane Sherburn and James Aung and Jun Shern Chan and Leon Maksin and Rachel Dias and Evan Mays and Benjamin Kinsella and Wyatt Thompson and Johannes Heidecke and Amelia Glaese and Tejal Patwardhan},
booktitle={Forty-second International Conference on Machine Learning},
year={2025},
url={https://openreview.net/forum?id=xF5PuTLPbn}
}

@article{tian2024scicode,
  title={Scicode: A research coding benchmark curated by scientists},
  author={Tian, Minyang and Gao, Luyu and Zhang, Shizhuo D and Chen, Xinan and Fan, Cunwei and Guo, Xuefei and Haas, Roland and Ji, Pan and Krongchon, Kittithat and Li, Yao and others},
  journal={Advances in Neural Information Processing Systems},
  volume={37},
  pages={30624--30650},
  year={2024}
}

@article{shen2024vlstem,
  title={Measuring vision-language stem skills of neural models},
  author={Shen, Jianhao and Yuan, Ye and Mirzoyan, Srbuhi and Zhang, Ming and Wang, Chenguang},
  journal={arXiv preprint arXiv:2402.17205},
  year={2024}
}

@article{zaki2023mascqa,
  title={Mascqa: A question answering dataset for investigating materials science knowledge of large language models},
  author={Zaki, Mohd and Krishnan, NM and others},
  journal={arXiv preprint arXiv:2308.09115},
  year={2023}
}

@software{UK_AI_Security_Institute_Inspect_AI_Framework_2024,
  author = {AI Security Institute, UK},
  title = {Inspect {AI:} {Framework} for {Large} {Language} {Model}
    {Evaluations}},
  date = {2024-05},
  url = {https://github.com/UKGovernmentBEIS/inspect_ai},
  langid = {en}
}

@article{bieganowski2025putnamlike,
  title={Putnam-like dataset summary: LLMs as mathematical competition contestants},
  author={Bieganowski, Bartosz and Strzelecki, Daniel and Skiba, Robert and Topolewski, Mateusz},
  journal={arXiv preprint arXiv:2509.24827},
  year={2025}
}

@article{deltoroiniesta2016inversion,
  author  = {del Toro Iniesta, Jose Carlos and Ruiz Cobo, Basilio},
  title   = {Inversion of the radiative transfer equation for polarized light},
  journal = {Living Reviews in Solar Physics},
  volume  = {13},
  pages   = {4},
  year    = {2016},
  doi     = {10.1007/s41116-016-0005-2}
}

@article{mistryukova2023stokes,
  author  = {Mistryukova, Lukia and Plotnikov, Andrey and Khizhik, Aleksandr and Knyazeva, Irina and Hushchyn, Mikhail and Derkach, Denis},
  title   = {Stokes inversion techniques with neural networks: Analysis of uncertainty in parameter estimation},
  journal = {Solar Physics},
  volume  = {298},
  pages   = {98},
  year    = {2023},
  doi     = {10.1007/s11207-023-02189-4}
}

@inproceedings{kan2022brain,
  author    = {Kan, Xuan and Dai, Wei and Cui, Hejie and Zhang, Zilong and Guo, Ying and Yang, Carl},
  title     = {Brain network transformer},
  booktitle = {Advances in Neural Information Processing Systems},
  volume    = {35},
  year      = {2022}
}

@article{popov2024meanmlp,
  author  = {Popov, Pavel and Mahmood, Usman and Fu, Zening and Yang, Carl and Calhoun, Vince and Plis, Sergey},
  title   = {A simple but tough-to-beat baseline for {fMRI} time-series classification},
  journal = {NeuroImage},
  volume  = {303},
  pages   = {120909},
  year    = {2024},
  doi     = {10.1016/j.neuroimage.2024.120909}
}

@article{sanzleon2015virtual,
  author  = {Sanz-Leon, Paula and Knock, Stuart A. and Spiegler, Andreas and Jirsa, Viktor K.},
  title   = {Mathematical framework for large-scale brain network modeling in {The Virtual Brain}},
  journal = {NeuroImage},
  volume  = {111},
  pages   = {385--430},
  year    = {2015},
  doi     = {10.1016/j.neuroimage.2015.06.016}
}

@article{deco2014local,
  author  = {Deco, Gustavo and Ponce-Alvarez, Adr{\'i}an and Hagmann, Patric and Romani, Gian Luca and Mantini, Dante and Corbetta, Maurizio},
  title   = {How local excitation--inhibition ratio impacts the whole brain dynamics},
  journal = {Journal of Neuroscience},
  volume  = {34},
  number  = {23},
  pages   = {7886--7898},
  year    = {2014},
  doi     = {10.1523/JNEUROSCI.5068-13.2014}
}

@article{wilsoncowan1972excitatory,
  author  = {Wilson, Hugh R. and Cowan, Jack D.},
  title   = {Excitatory and inhibitory interactions in localized populations of model neurons},
  journal = {Biophysical Journal},
  volume  = {12},
  number  = {1},
  pages   = {1--24},
  year    = {1972},
  doi     = {10.1016/S0006-3495(72)86068-5}
}

@article{papadopoulos2020relations,
  author  = {Papadopoulos, Lia and Lynn, Christopher W. and Battaglia, Demian and Bassett, Danielle S.},
  title   = {Relations between large-scale brain connectivity and effects of regional stimulation depend on collective dynamical state},
  journal = {PLoS Computational Biology},
  volume  = {16},
  number  = {9},
  pages   = {e1008144},
  year    = {2020},
  doi     = {10.1371/journal.pcbi.1008144}
}

@article{hansen2022mapping,
  author  = {Hansen, Justine Y. and Shafiei, Golia and Markello, Ross D. and others},
  title   = {Mapping neurotransmitter systems to the structural and functional organization of the human neocortex},
  journal = {Nature Neuroscience},
  volume  = {25},
  pages   = {1569--1581},
  year    = {2022},
  doi     = {10.1038/s41593-022-01186-3}
}

@article{handwerker2004variation,
  author  = {Handwerker, Daniel A. and Ollinger, John M. and D'Esposito, Mark},
  title   = {Variation of {BOLD} hemodynamic responses across subjects and brain regions and their effects on statistical analyses},
  journal = {NeuroImage},
  volume  = {21},
  number  = {3},
  pages   = {1639--1651},
  year    = {2004},
  doi     = {10.1016/j.neuroimage.2003.11.031}
}

@article{glover1999deconvolution,
  author  = {Glover, Gary H.},
  title   = {Deconvolution of impulse response in event-related {BOLD} {fMRI}},
  journal = {NeuroImage},
  volume  = {9},
  number  = {5},
  pages   = {416--429},
  year    = {1999},
  doi     = {10.1006/nimg.1998.0419}
}

@article{friston2000nonlinear,
  author  = {Friston, Karl J. and Mechelli, Andrea and Turner, Robert and Price, Cathy J.},
  title   = {Nonlinear responses in {fMRI}: The {Balloon} model, {Volterra} kernels, and other hemodynamics},
  journal = {NeuroImage},
  volume  = {12},
  number  = {4},
  pages   = {466--477},
  year    = {2000},
  doi     = {10.1006/nimg.2000.0630}
}

@article{wu2013blind,
  author  = {Wu, Guo-Rong and Liao, Wei and Stramaglia, Sebastiano and Ding, Ju-Rong and Chen, Huafu and Marinazzo, Daniele},
  title   = {A blind deconvolution approach to recover effective connectivity brain networks from resting state {fMRI} data},
  journal = {Medical Image Analysis},
  volume  = {17},
  number  = {3},
  pages   = {365--374},
  year    = {2013},
  doi     = {10.1016/j.media.2013.01.003}
}

@article{wei2025agentic,
  title   = {From {AI} for Science to Agentic Science: A Survey on Autonomous Scientific Discovery},
  author  = {Wei, Jiaqi and others},
  journal = {arXiv preprint arXiv:2508.14111},
  year    = {2025}
}

@article{qiu2025supernova,
  title   = {Physics Supernova: {AI} Agent Matches Elite Gold Medalists at {IPhO} 2025},
  author  = {Qiu, Jiahao and others},
  journal = {arXiv preprint arXiv:2509.01659},
  year    = {2025}
}

@article{mitchener2025kosmos,
  title   = {Kosmos: An {AI} Scientist for Autonomous Discovery},
  author  = {Mitchener, Ludovico and others},
  journal = {arXiv preprint arXiv:2511.02824},
  year    = {2025}
}

@article{demianenko2023understanding,
  title={Understanding of the properties of neural network approaches for transient light curve approximations},
  author={Demianenko, Mariia and Malanchev, Konstantin and Samorodova, Ekaterina and Sysak, Mikhail and Shiriaev, Aleksandr and Derkach, Denis and Hushchyn, Mikhail},
  journal={Astronomy \& Astrophysics},
  volume={677},
  pages={A16},
  year={2023},
  publisher={EDP Sciences}
}

@article{aarrestad2021dark,
  title={The dark machines anomaly score challenge: benchmark data and model independent event classification for the large hadron collider},
  author={Aarrestad, Thea and van Beekveld, Melissa and Bona, Marcella and Boveia, Antonio and Caron, Sascha and Davies, Joe and De Simone, Andrea and Doglioni, Caterina and Duarte, Javier M and Farbin, Amir and others},
  journal={arXiv preprint arXiv:2105.14027},
  year={2021}
}

@article{bhimji2026fair,
  title={FAIR universe higgsML uncertainty dataset and competition},
  author={Bhimji, Wahid and Chakkappai, Ragansu and Chang, Po-Wen and Chou, Yuan-Tang and Diefenbacher, Sascha and Dudley, Jordan and Elsharkawy, Ibrahim and Farrell, Steven and Ghosh, Aishik and Giordano, Cristina and others},
  journal={Advances in Neural Information Processing Systems},
  volume={38},
  year={2026}
}

@article{kasieczka2021lhc,
  title={The LHC Olympics 2020 a community challenge for anomaly detection in high energy physics},
  author={Kasieczka, Gregor and Nachman, Benjamin and Shih, David and Amram, Oz and Andreassen, Anders and Benkendorfer, Kees and Bortolato, Blaz and Brooijmans, Gustaaf and Canelli, Florencia and Collins, Jack H and others},
  journal={Reports on progress in physics},
  volume={84},
  number={12},
  pages={124201},
  year={2021},
  publisher={IOP Publishing}
}

@inproceedings{qu2022particle,
  title={Particle transformer for jet tagging},
  author={Qu, Huilin and Li, Congqiao and Qian, Sitian},
  booktitle={International Conference on Machine Learning},
  pages={18281--18292},
  year={2022},
  organization={PMLR}
}
